\documentclass[10pt,twocolumn,letterpaper]{article}

\usepackage[pagenumbers]{cvpr} 

\usepackage[dvipsnames]{xcolor}

\usepackage{algorithm}
\usepackage[noend]{algpseudocode}

\usepackage{mathtools, nccmath}
\usepackage{tabularx}
\usepackage{caption}
\usepackage{multirow}

\usepackage{amsmath}
\usepackage{booktabs}
\usepackage{siunitx}

\usepackage[symbol,hang,flushmargin]{footmisc}

\definecolor{cvprblue}{rgb}{0.21,0.49,0.74}
\usepackage[pagebackref,breaklinks,colorlinks,citecolor=cvprblue]{hyperref}

\let\oldtwocolumn\twocolumn
\renewcommand\twocolumn[1][]{%
    \oldtwocolumn[{#1}{
    \begin{center}
 \includegraphics[width=\linewidth]{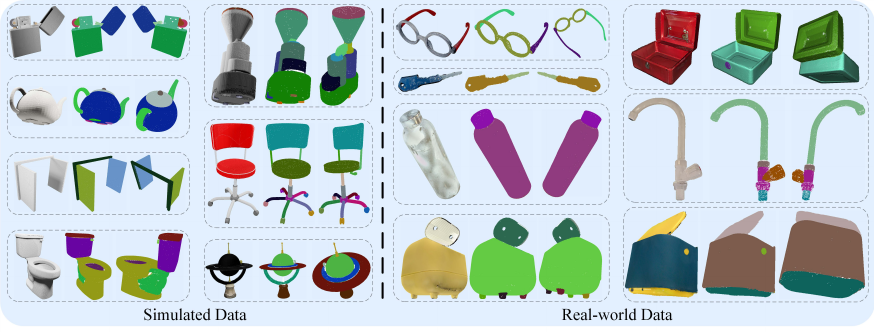}
           \captionof{figure}{We propose ZeroPS, a novel zero-shot 3D part segmentation pipeline by leveraging pretrained foundation models, SAM \cite{kirillov2023segment} and GLIP \cite{li2022grounded}, without training or fine-tuning. The figure shows our unlabeled segmentation results. Left: PartNetE's simulated data. Right: AKBSeg's real-world data. In each example, the input 3D object and output two visualizations are shown in turn. Please zoom in to see accurate 3D segmentation boundaries. ZeroPS also supports instance segmentation. Please refer to \cref{fig:comparewithpartslip} for the qualitative comparison.} 
           \label{fig:poster}
        \end{center}
    }]
}

\title{ZeroPS: High-quality Cross-modal Knowledge Transfer for Zero-Shot 3D Part Segmentation}

\author{Yuheng Xue$^{1}$ \qquad
    Nenglun Chen$^{1}$ \qquad
    Jun Liu$^{2}$ \qquad 
    Wenyun Sun$^{1\diamond}$\\
    $^{1}$Nanjing University of Information Science and Technology, China\\
    $^{2}$Lancaster University, UK \\
    {\tt\small \{yuhengxue,chennenglun,wenyunsun\}@nuist.edu.cn}  \ 
    {\tt\small \{j.liu81\}@lancaster.ac.uk} 
}

\begin{document} 

\setlength{\textfloatsep}{5pt plus 0.5pt minus 0.5pt}
\setlength{\dblfloatsep}{4.5pt plus 0.5pt minus 0.5pt}
\setlength{\dbltextfloatsep}{4pt plus 0.5pt minus 0.5pt}

\setlength{\parskip}{-1pt}
\setlength{\belowdisplayshortskip}{4pt}
\setlength{\belowdisplayskip}{4pt} 
\setlength{\abovedisplayshortskip}{4pt}
\setlength{\abovedisplayskip}{4pt}  

\maketitle

\setlength{\skip\footins}{0.5em} 
\newcommand\blfootnote[1]{%
  \begingroup
  \renewcommand\thefootnote{}\footnote{#1}%
  \addtocounter{footnote}{-1}%
  \endgroup
}

\blfootnote{$\diamond$: Corresponding author}

\begin{abstract}
Zero-shot 3D part segmentation is a challenging and fundamental task. 
In this work, we propose a novel pipeline, ZeroPS, which achieves high-quality knowledge transfer from 2D pretrained foundation models (FMs), SAM and GLIP, to 3D object point clouds. 
We aim to explore the natural relationship between multi-view correspondence and the FMs' prompt mechanism and build bridges on it. 
In ZeroPS, the relationship manifests as follows: 1) lifting 2D to 3D by leveraging co-viewed regions and SAM's prompt mechanism, 2) relating 1D classes to 3D parts by leveraging 2D-3D view projection and GLIP's prompt mechanism, and 3) enhancing prediction performance by leveraging multi-view observations. 
Extensive evaluations on the PartNetE and AKBSeg benchmarks demonstrate that ZeroPS significantly outperforms the SOTA method across zero-shot unlabeled and instance segmentation tasks. 
ZeroPS does not require additional training or fine-tuning for the FMs. 
ZeroPS applies to both simulated and real-world data. 
It is hardly affected by domain shift. 
The project page is available at \url{https://luis2088.github.io/ZeroPS_page/}.
\end{abstract}

\vspace{-2em}
\section{Introduction} 
3D part segmentation is a crucial task in computer vision and computer graphics, leading to various applications such as robotics, shape editing, and AR/VR \cite{liu2022frame,aleotti20123d,xu2022unsupervised,lin2022neuform,umam2022point}. Due to the scarcity of 3D training data, recent research efforts focus on leveraging knowledge from foundation models (FMs) in other modalities (\eg, text or images) to design a zero-shot manner, i.e., zero-shot 3D part segmentation. 
During inference, zero-shot 3D part segmentation requires not only making predictions on unseen data and classes but also ensuring accurate 3D segmentation. 
Though challenging, this task aligns with practical scenarios, like accurately segmenting a 3D object in an unfamiliar environment. 

In this work, we propose a novel pipeline, ZeroPS, which achieves high-quality knowledge transfer from 2D pretrained FMs, SAM \cite{kirillov2023segment} and GLIP \cite{li2022grounded}, to 3D object point clouds. 
We aim to explore the natural relationship between multi-view correspondence and the FMs' prompt mechanism and build bridges on it. 
The following two subsections describe the manifestations of the relationship. 

\begin{figure*}[!tb]
\centering
  \includegraphics[width=\linewidth]{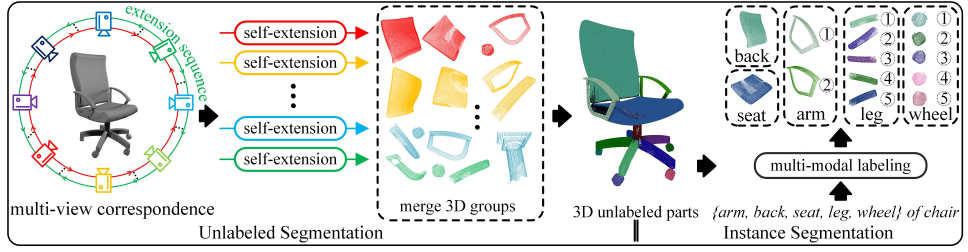}
  \caption{
  Overview of the proposed pipeline ZeroPS. 
  First, in the unlabeled segmentation phase, the input 3D object is segmented into unlabeled parts. The self-extension (See \cref{fig:self-extension}) component can extend 2D segmentation from a single viewpoint to 3D segmentation (3D groups), by using a predefined extension sequence starting from that viewpoint. For example, the red cue on the left side of the figure illustrates this process. Second, in the instance segmentation phase, given a text prompt, the multi-modal labeling (See \cref{fig:mml}) component assigns an instance label to each 3D unlabeled part. 
  }
  \label{fig:pipeline}
\end{figure*}

Intuitively, for a 3D object, we can obtain 2D groups by leveraging SAM to segment 2D images from different viewpoints. 
By back-projecting these groups into 3D and merging them, we can obtain 3D unlabeled parts. 
However, an important insight is that there exists a natural relationship between co-viewed regions and SAM's prompt mechanism. 
For any group in 3D, the visible portion of the group in adjacent viewpoints can be used as SAM's prompt to further extend it. 
By leveraging other viewpoints to continuously extend the 2D segmentation results, they will gradually become more complete in 3D. 
Therefore, as shown in \cref{fig:pipeline,fig:self-extension}, we design a component self-extension, which obtains 2D groups and extends each group from 2D to 3D. \textbf{Self-extension leverages the natural relationship between co-viewed regions and SAM’s prompt mechanism to lift 2D to 3D in a training-free manner}. 

To assign an instance label to each 3D unlabeled part, we integrate the GLIP model. 
As shown in \cref{fig:mml}, given a text prompt containing part classes, GLIP predicts many 2D bounding boxes. 
Since 2D boxes and 3D parts are not in the same space, we propose a two-dimensional checking mechanism (TDCM) to vote each 2D box to the best-matched 3D part, yielding a Vote Matrix. We select the highest vote in each column (part), thereby assigning an instance label to each 3D part. 
\textbf{TDCM leverages the natural relationship between 2D-3D view projection and GLIP's prompt mechanism (1D-2D) to relate 1D to 3D in a training-free manner.} 
To enhance the accuracy of label assignment for 3D parts, we propose a Class Non-highest Vote Penalty (CNVP) function to refine the Vote Matrix. 
Since GLIP inevitably produces incorrect predictions, the Vote Matrix exhibits certain unfairness (See \cref{sec:multi-modelabeling} for details). 
An insight is that in each row (class) of the Vote Matrix, the highest vote represents GLIP's prediction for that class across as many views as possible, indicating most likely to be the correct prediction. 
CNVP penalizes other votes by using the highest vote in each row (class), yielding a refined version of the Vote Matrix. 
\textbf{CNVP leverages multi-view observations, which allows it to enhance prediction performance while retaining GLIP's zero-shot generalization in a training-free manner.} 

In the experiment, we conduct extensive evaluations on the public PartNetE \cite{liu2023partslip} benchmark. 
Our method outperforms the SOTA method by a large margin across unlabeled and instance segmentation tasks. 
Our method further narrows the gap between zero-shot and 3D fully supervised counterparts. 
Ablation studies demonstrate that both self-extension and CNVP contribute significantly to performance improvements. 
To better evaluate the generalization of all zero-shot methods concerning \textit{unseen data, unseen classes, and hyperparameters}, we propose an AKBSeg benchmark from the existing AKB-48 \cite{liu2022akb} dataset. 
Retaining the default configurations from the evaluation on PartNetE, all zero-shot methods are re-evaluated on the AKBSeg benchmark. 
Our method continues to outperform the SOTA method by a large margin. Overall, the main contributions of our paper include: 
\begin{itemize} 
	\item A novel zero-shot 3D part segmentation pipeline by leveraging pretrained foundation models, SAM and GLIP, without training or fine-tuning, is based only on multi-view correspondence and the foundation models' prompt mechanism, allowing it to demonstrate superior segmentation performance while preserving the zero-shot generalization from the pretrained foundation models. 
        \item Three training-free manners: 1) self-extension lifts 2D to 3D by leveraging co-viewed regions and SAM’s prompt mechanism; 2) TDCM relates 1D classes to 3D parts by leveraging 2D-3D view projection and GLIP's prompt mechanism; and 3) CNVP enhances prediction performance by leveraging multi-view observations. 
        \item Our method achieves better zero-shot generalization and segmentation performance than the SOTA method. 
\end{itemize} 

\section{Related Work} 
\subsection{Supervised 3D Segmentation} 
Most methods \cite{qi2017pointnet,qi2017pointnet++,wang2019dynamic,qian2022pointnext,thomas2019kpconv,vu2022softgroup,zhao2021pointtransformer} are fully supervised training on 3D datasets. 
These works focus on the design of network architectures to learn better 3D representations. 
The classical PointNet \cite{qi2017pointnet} considers the data structure of 3D points. 
Subsequent works \cite{yi2019gspn,wang2018sgpn,yu2019partnet,lai2022stratified,liu2020self,luo2020learning,zhang2021point,cao2023mopa,liu2022autogpart,wu2024point,wu2022point} introduce ideas from the common deep learning field, such as transformer \cite{vaswani2017attention}, unet \cite{ronneberger2015u}, graph CNN \cite{kipf2016semi,defferrard2016convolutional}, rpn \cite{ren2015faster}, etc. 
However, the 3D datasets \cite{yi2016scalable,mo2019partnet,geng2023gapartnet} are several orders of magnitude smaller than the image datasets \cite{russakovsky2015imagenet}, but the complexity of 3D data is higher than images. 
Therefore, many works make up for the defects of insufficient 3D data through different training strategies, such as weak supervision \cite{chibane2022box2mask,xu2020weakly,wang2022ikea,koo2022partglot}, 
self-supervision \cite{yu2022pointbert,gadelha2020label,liu2022masked,pang2022masked} or few-shot learning \cite{sharma2020self,wang2020few,sharma2022mvdecor,an2024rethinking,kim2025partstad,zhou2023partslip++,kareem2025paris3d}. 
In this work, we contrast zero-shot methods and 3D fully supervised counterparts on quantitative metrics. 

\subsection{2D Foundation Models (FMs)} 
Recently, 2D FMs trained on large-scale datasets have demonstrated impressive zero-shot generalization. 
Another notable characteristic is the rich prompt mechanisms that enable these 2D FMs to establish connections across different modalities. 
Using free-form text prompts, CLIP \cite{radford2021learning} generates pixel-level predictions for a given image. 
A series of zero-shot 2D detectors, such as GLIP \cite{li2022grounded}, GDINO \cite{ren2024grounding,liu2023grounding}, and Yolo-world \cite{cheng2024yolo}, can output prediction bounding boxes for target objects based on given the text template prompt (e.g., `{arm, back, seat, wheel, leg} of chair'). 
Segment Anything (SAM) \cite{kirillov2023segment} is an FM for zero-shot 2D segmentation. 
Given an image, SAM output instance masks at three different granularities (whole, part, and subpart) based on point or box prompts. 
Due to SAM's groundbreaking impact in 2D segmentation, many SAM-like models enhance various aspects of SAM, such as quality \cite{ke2024segment,chen2024robustsam} and efficiency \cite{xiong2023efficientsam,zhang2023faster,zhao2023fast}. 
For more works about FMs, please refer to this review \cite{awais2023foundational}. 
This work leverages 1) the 2D instance-level segmentation capability and point prompt mechanism of a pretrained SAM model and 2) the 2D instance-level detection capability and text template prompt mechanism of a pretrained GLIP model. 

\subsection{Zero-shot 3D Part Segmentation} 
Except for: 1) the lack of large-scale 3D training data and 2) the robust zero-shot generalization of 2D FMs, another factor is 3) the 2D-3D mapping can be established through view projection and back-projection. 
These three factors drive recent research into exploring how to leverage 2D FMs to perform zero-shot 3D part segmentation. 

Most existing works directly transfer knowledge from 2D FMs through multi-view 2D-3D mapping. 
PointCLIP V2 \cite{zhu2023pointclip} proposes a realistic projection technique to enhance CLIP's visual encoder. 
It enables PointCLIP V2 to segment 3D sparse object point clouds. 
GeoZe \cite{mei2024geometrically} considers the intrinsic geometric information of 3D objects and proposes a training-free geometry-driven aggregation strategy. 
PartSLIP \cite{liu2023partslip} uses the predicted bounding boxes from GLIP, to determine each initial superpoint's semantics. 
A well-designed grouping module partitions all semantic superpoints into 3D instance parts. 
Satr \cite{Abdelreheem_2023_ICCV} also utilizes GLIP, but while PartSLIP focuses on point cloud segmentation, Satr is oriented toward mesh segmentation. 
Unlike these methods, PartDistill \cite{umam2024partdistill} introduces a bi-directional distillation framework, distilling 2D knowledge from CLIP or GLIP into a 3D student network, fully leveraging unlabeled 3D data for end-to-end training. 
However, limited by predefined output classes, the 3D network can predict unseen data but not unseen classes, with CLIP or GLIP's capability in this regard not retained in 3D. 
Our work explores new ideas for directly transferring knowledge. 

The existing work mostly focuses on semantic segmentation \cite{zhu2023pointclip,mei2024geometrically,Abdelreheem_2023_ICCV,umam2024partdistill,thai20253}, with only PartSLIP \cite{liu2023partslip} capable of performing instance segmentation. 
To address the gap in instance segmentation methods, this work continues to explore new ideas for zero-shot instance segmentation. 

Another parallel direction investigates zero-shot methods for scene segmentation \cite{yang2023sam3d,xu2023sampro3d,huang2025segment3d,yan2024maskclustering,yin2024sai3d,takmaz2023openmask3d,nguyen2024open3dis}. 
\begin{figure*}[!tb]
\includegraphics[width=\linewidth]{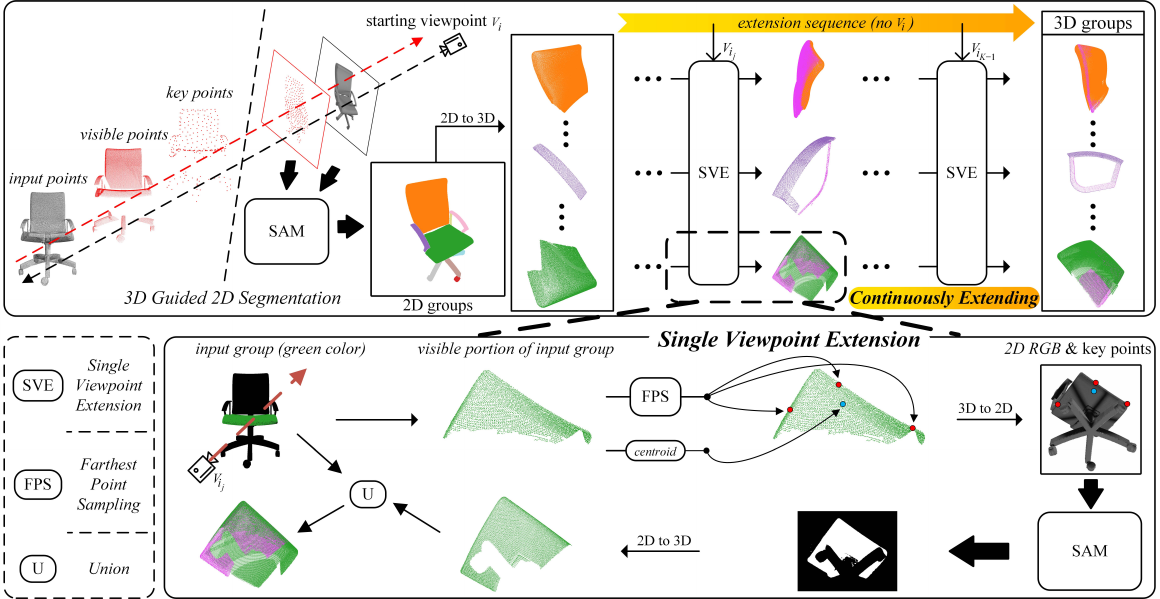}
\caption{The overall structure of self-extension (top subfigure). 
Given an extension sequence $S_i = [V_i, V_{i_1}, V_{i_2}, \ldots, V_{i_j}, \ldots, V_{i_{k-1}}]$, self-extension aims to obtain 2D groups from starting viewpoint $V_i$ and extends each group from 2D to 3D by the remaining viewpoints. 
Specifically, for the starting viewpoint $V_i$, self-extension utilizes 3D key points to guide SAM to segment the 2D image. As the segmented 2D groups originate from 2D segmentation results, self-extension continuously extends these groups to 3D segmentation results (3D groups) by SVE (Single Viewpoint Extension). 
During continuously extending, the remaining viewpoints in $S_i$, $[V_{i_1}, V_{i_2}, \ldots, V_{i_j}, \ldots, V_{i_{k-1}}]$, are iterated. 
At each iteration, inputting the current viewpoint and each group, SVE extends each input group. As an example, a detailed process of how SVE extends a single group is provided in the bottom right subfigure.} 
\label{fig:self-extension}
\end{figure*} 

\section{Proposed Method: ZeroPS} 
\subsection{Overview} 
Given a 3D object point cloud, this work aims to utilize SAM and GLIP to perform two types of segmentation: unlabeled and instance segmentation. 
The overall pipeline (See \cref{fig:pipeline}) is divided into two phases. 
In the unlabeled part segmentation phase, we first define the following operators by multi-view correspondence (See \cref{sec:Multi-view Correspondence}): 1) obtaining the extension sequence $S_i$ starting from any viewpoint $V_i$; 2) calculating the forward- and back-projection between any viewpoint $V_i$ and 3D space. 
Then, each self-extension (See \cref{sec:Self-extension}) component inputs an extension sequence $S_i$ and outputs 3D groups. 
Next, we merge all 3D groups (See \cref{sec:Merge3DGroups}) by a merging algorithm and get 3D unlabeled parts. 
In the instance segmentation phase, the multi-model labeling (See \cref{sec:multi-modelabeling}) component assigns an instance label to each 3D unlabeled part based on a text prompt. 

\subsection{Multi-view Correspondence} 
\label{sec:Multi-view Correspondence} 
In 3D space, given an object point cloud $Q^{3D}\in\mathbb{R}^{N\times6}$ as input, where $N$ represents the number of points, each point includes a position $\{x,y,z\}$ and color $\{r,g,b\}$. 
We arrange $K$ viewpoints relatively uniformly around $Q^{3D}$. 
It can be referred to Table S1 in Supplementary. 
We use the notation $V_i$ $(i = 1, 2, \ldots, K)$ to name each viewpoint $V$. 
Then, we perform point cloud rendering of \( Q^{3D} \) from each viewpoint \( V_i \). 
The output of each $V_i$ consists of a 2D RGB image denoted as $I_i$ with shape $(H \times W \times 3)$ and a point cloud index matrix denoted as $P_i$ with shape $(H \times W \times 1)$, where each element at matching locations in $I_i$ and $P_i$ respectively represent the 3D position and color of the same point. 
Now, for any point of $Q^{3D}$ in 3D space, we can easily find its position in the pixel coordinate system of $V_i$, or vice versa. 

\textbf{Extension Sequence.} We construct an undirected, unweighted graph using all \( K \) viewpoints surrounding the 3D object \( Q^{3D} \). 
In this graph, each node represents a viewpoint, and edges are established between adjacent viewpoints based on their spatial arrangement. 
Starting from any viewpoint \( V_i \), we can perform a breadth-first search algorithm on the graph, resulting in a sequence referred to as the extension sequence, denoted as \( S_i = [V_i, V_{i_1}, V_{i_2}, \ldots, V_{i_j}, \ldots, V_{i_{K-1}}] \). 
The extension sequence ensures that at any iteration, there is a co-viewed region between the current viewpoint and the previously iterated viewpoints. 
This allows, during continuously extending, Single Viewpoint Extension (SVE) to obtain the visible portion of the input group from the input viewpoint. 

\textbf{Bi-directional Projection (BiP)}. To facilitate the mapping of a subset of $Q^{3D}$ between 2D (any $V_i$) and 3D space, we denote forward- and back-projection simply as BiP as follows: 
\begin{gather}
  X^{3D}=BiP(X^{2D},P_{i}),
  \label{eq:PT1}\\
  X^{2D}=BiP(X^{3D},P_{i}),
  \label{eq:PT2}
\end{gather} 
where $X^{3D}$ indicates a subset of $Q^{3D}$ and $X^{2D}$ indicates the subset of 2D coordinates of the pixel coordinate system of $V_i$. 
More generally, BiP can also in parallel process multiple subsets in the same viewpoint. 

\subsection{Self-extension} 
\label{sec:Self-extension} 
Self-extension aims to obtain 2D groups and extends each group from 2D to 3D. 
The overall structure of self-extension is illustrated in \cref{fig:self-extension}. 
Given an extension sequence $S_i = [V_i, V_{i_1}, V_{i_2}, \ldots, V_{i_j}, \ldots, V_{i_{k-1}}]$, for the starting viewpoint $V_i$, self-extension utilizes 3D key points to guide SAM to segment the 2D image. 
As the segmented 2D groups originate from 2D segmentation results, self-extension continuously extends these groups to 3D segmentation by SVE. 

\textbf{3D Guided 2D Segmentation.} 
As shown in the top subfigure of \cref{fig:self-extension}, to obtain 2D groups, all key points and 2D RGB image $I_i$ in $V_i$ are fed into SAM.
An automatic segmentation setting is then performed. 
The overall process can be formulated as follows: 
\begin{gather}
    KPoints_{V_i}^{2D}=BiP(FPS(Q_{V_i}^{3D}),P_{i}),
    \label{eq:KP2D}\\
     \{G_{1}^{2D},\ldots,G_{n}^{2D}\}=SAM(I_{i},KPoints_{V_i}^{2D}),
     \label{eq:G2Ds}
\end{gather}
where $Q_{V_i}^{3D}$ indicates the visible points of $Q^{3D}$ in $V_i$, $KPoints^{2D}$ indicates key points, $\{G_{1}^{2D},\ldots,G_{n}^{2D}\}$ indicates $n$ 2D groups, and $FPS$ indicates Farthest Point Sampling. 
These 2D groups are back-projected into 3D space:
\begin{equation}
\{G_{1}^{3D},\ldots,G_{n}^{3D}\}= BiP(\{G_{1}^{2D},\ldots,G_{n}^{2D}\},P_i).
 \label{eq:G3Dss}
\end{equation} 

\textbf{Single Viewpoint Extension (SVE).} Before continuously extending, a Single Viewpoint Extension (SVE) operator needs to be defined. 
Given a viewpoint, SVE can extend the input group. 
For a group in 3D, we observe a natural relationship between the co-viewed region and SAM’s prompt mechanism. 
As shown in the bottom right subfigure of \cref{fig:self-extension}, from $V_{i_j}$, `we observe' a portion of the input group (green color), as there is a co-viewed region between $V_{i_j}$ and $\{V_{i_1}, V_{i_2}, \ldots, V_{i_{j-1}}\}$ (See \cref{sec:Multi-view Correspondence}, `Extension Sequence'). 
SVE obtains the visible portion of the input group from $V_{i_j}$. 
Then, SVE feeds both $I_{i_j}$ as 2D RGB and the key points as prompt into SAM and performs inference. 
Finally, SVE obtains the union of the mask and the input group. 
The input group is extended to more points with the same semantics. 
The overall process can be formulated as: 
\begin{gather}
    KPoints_{V_{i_j}}^{2D}=BiP(FPS(G_{V_{i_j}}^{3D})\cup CC(G_{V_{i_j}}^{3D}),P_{i_j}),
  \label{eq:KP2D1}\\
   Mask = SAM(I_{i_j},KPoints_{V_{i_j}}^{2D}),
  \label{eq:mask}\\
  G^{3D} \leftarrow G^{3D}\cup BiP(Mask,P_{i_j}),
  \label{eq:extend}
\end{gather}
where $G_{V_{i_j}}^{3D}$ indicates the visible portion of the input group from $V_{i_j}$, $KPoints_{V_{i_j}}^{2D}$ indicates key points in $V_{i_j}$, $\leftarrow$ indicates set extension and $CC$ indicates the calculation of the point closest to the centroid. 
We propose the Single Viewpoint Extension (SVE) with input a $G^{3D}$ and $V$: 
\begin{equation} 
  G^{3D} \leftarrow SVE(G^{3D},V),
  \label{eq:KP2D2}
\end{equation}
where SVE is utilized to extend the $G^{3D}$ from $V$. 
For SVE's input, $G^{3D}$ needs to be within the visual range of viewpoint $V$. 
Otherwise, it will not be extended (remaining unchanged). 
More generally, SVE can in parallel extend a set of groups in the same viewpoint. 

\textbf{Continuously Extending.} To continuously extend each group of \cref{eq:G3Dss}, we iterate over the remaining viewpoints of $S_i$, $[V_{i_1}, V_{i_2}, \ldots, V_{i_j}, \ldots, V_{i_{K-1}}]$, by SVE:  
\begin{equation}
\begin{aligned}\label{eq:SS2ES}
& \{G_{1}^{3D},\ldots,G_{n}^{3D}\} \leftarrow SVE(\{G_{1}^{3D},\ldots,G_{n}^{3D}\}, V_{i_1}) \\
& \{G_{1}^{3D},\ldots,G_{n}^{3D}\} \leftarrow SVE(\{G_{1}^{3D},\ldots,G_{n}^{3D}\}, V_{i_2}) \\
& \phantom{\{G_{1}^{3D},\ldots,G_{n}^{3D}\} \leftarrow} \vdots \phantom{= SVE(\{G_{1}^{3D},\ldots,G_{n}^{3D}\}, V_{i_{K-1}})} \\
& \{G_{1}^{3D},\ldots,G_{n}^{3D}\} \leftarrow SVE(\{G_{1}^{3D},\ldots,G_{n}^{3D}\}, V_{i_{K-1}}).
\end{aligned}
\end{equation} 
Finally, each group in $\{G_{1}^{3D},\ldots,G_{n}^{3D}\}$ is extended from 2D (the starting viewpoint $V_i$) to 3D (all viewpoints). 
In summary, the self-extension can be represented as:
\begin{equation}
    \{G_{1}^{3D},\ldots,G_{n}^{3D}\}=SE(S_{i}),
    \label{eq:esvi}
\end{equation}
where $S_{i}$ indicates an extension sequence starting from $V_i$, $SE$ indicates self-extension component and $\{G_{1}^{3D},\ldots,G_{n}^{3D}\}$ indicates a set of 3D groups resulting from $SE(S_{i})$ starting from $V_i$. 

\subsection{Merging 3D Groups} 
\label{sec:Merge3DGroups} 
To get 3D unlabeled parts, a merging algorithm is employed to merge 3D groups, which are the output of all self-extensions (See \cref{fig:pipeline}). 
The algorithm depends on a merging threshold T. The pseudocode and detailed explanation are in the supplementary materials. 

\subsection{Multi-model Labeling} 
\label{sec:multi-modelabeling} 
\begin{figure}[!tb]
    \centering
   \includegraphics[width=\linewidth]{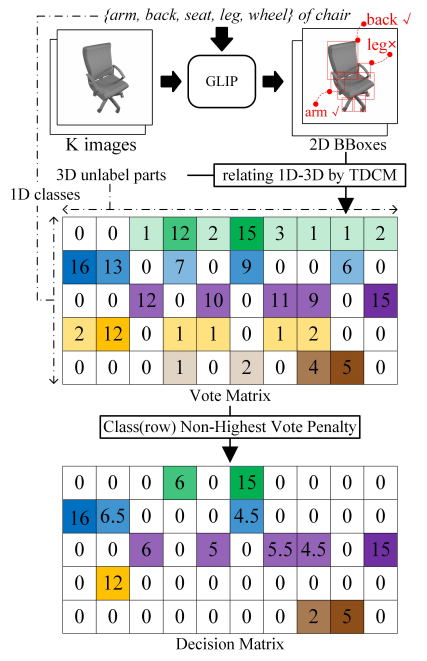}
   \caption{The overall structure of multi-modal labeling.}
   \label{fig:mml}
\end{figure}

Multi-model labeling aims to assign an instance label to each 3D unlabeled part. 
The main idea is shown in \cref{fig:mml}. 
To get lots of 2D bounding boxes with instance labels, a text prompt containing part classes and K images (from all viewpoints) are fed into GLIP. 
Then, we vote each box to the best-matched 3D part and obtain a Vote Matrix that relates 1D classes (rows) to 3D parts (columns). 
Intuitively, we simply get the highest vote per column (part) and assign its class as a label to that part. 
However, we must face two problems: 1) How to vote each 2D bounding box to the best-matched 3D part, given that they are not in the same space; 
2) How to enhance the accuracy of label assignment for 3D parts, since GLIP inevitably produces incorrect predictions. 

\textbf{Two-dimensional Checking Mechanism (TDCM).} 
To vote each 2D predicted bounding box to the best-matched 3D part, we design a two-dimensional checking mechanism. 
Meanwhile, some unqualified boxes are discarded. 

In detail, for any 2D predicted bounding box $BB$, we perform the Intersection over Union (IoU) between the $F^{3D}$ and each 3D part $P^{3D}$ in $C=\{P_{1}^{3D},P_{2}^{3D},\ldots,P_{m_{2}}^{3D}\}$. 
Further, we let $P_s^{3D}$, with the Maximum IoU, be the best-matched 3D part in 3D space: 
\begin{equation}
   P_s^{3D} = \underset{P^{3D} \in C}{\arg\max} \frac{| F^{3D} \cap P^{3D}|}{| F^{3D}\cup P^{3D}|},
  \label{eq:ps3d}
\end{equation}
where the $F^{3D}$ indicates the 3D visible points inside the $BB$, and the $C$ indicates $m_{2}$ unlabeled parts $P^{3D}$. 
Meanwhile, we perform the IoU between the $BB$ and each $P^{box}$ in $C^{\prime}=\{P_{1}^{box},P_{2}^{box},\ldots,P_{m_{2}}^{box}\}$. 
Then we let $P_t^{box}$, with the Maximum IoU, be the best-matched 3D part in 2D space:
\begin{equation}
   P_t^{box} = \underset{P^{box} \in C^{\prime}}{\arg\max} \frac{| BB \cap P^{box}|}{| BB\cup P^{box}|},
  \label{eq:ps2d}
\end{equation}
where the $C^{\prime}$ indicates $m_{2}$ 2D bounding box $P^{box}$ of all 3D parts in the viewpoint where the $BB$ is located. 
Note that $P_{u}$ in the $C$ and $C^{\prime}$ denotes two states of the same 3D part, 3D point set and 2D bounding box, respectively. 
Finally, if $s=t$, the $BB$ is voted to $P_s^{3D}$. 
Otherwise, the $BB$ is discarded. 
In other words, it must guarantee that the best-matched part of the predicted bounding box in both 2D and 3D space is the same 3D part. 
Overall, TDCM leverages the natural relationship between 2D-3D view projection and GLIP’s prompt mechanism (1D-2D) to relate 1D classes to 3D parts into a Vote Matrix. 

\textbf{Class Non-highest Vote Penalty (CNVP).} 
To enhance the accuracy of label assignment for 3D parts, we propose a Class Non-highest Vote Penalty function. 

In fact, the 2D predicted bounding boxes produced by GLIP inevitably have incorrect labels (See top right of \cref{fig:mml}). 
When the highest vote per column (part) is directly obtained and its class assigned as a label to that 3D part, this leads to two kinds of unfairness: 1) For a specific column (part), the highest vote `wins' by only a narrow margin compared to other votes. 
For example, in the second column of the Vote Matrix in \cref{fig:mml}, `13' wins over `12' by just one vote; 
2) For different columns (parts), the gap between the highest votes is too large when their highest votes are in the same row (class). 
For example, compared to the highest vote `16', in the first column (part) of the Vote Matrix, the election of `6' in the penultimate column (part) is unreasonable. 
In this case, in the penultimate column (part), `5' is more trustworthy than `6', because `5' possesses the highest vote in the final row (class), while `6' does not even reach half of the highest vote in the second row (class). 

The unfairness in the Vote Matrix mentioned above needs improvement. 
In each row (class), the highest vote represents GLIP's prediction for that class across as many views as possible. 
This indicates that, compared to other votes, the highest vote is more likely to be the correct prediction, making it a reliable pivot in the Vote Matrix. 
Therefore, we use the highest vote per row (class) to penalize other votes through CNVP: 
\begin{equation}
       \begin{cases}
\alpha, & \text{if } \alpha / \alpha_{\mathrm{rm}} = 1 \\
\alpha / 2, & \text{if } 0.5 \leq \alpha / \alpha_{\mathrm{rm}} < 1 \\
0, & \text{if } 0 \leq \alpha / \alpha_{\mathrm{rm}} < 0.5
\end{cases},
  \label{eq:Penalty}
\end{equation}
where $\alpha$ indicates each element of the Vote Matrix, $\alpha_{rm}$ indicates the maximum value within the same row where $\alpha$ is located. 
CNVP results in a Decision Matrix, a refined version of the Vote Matrix. 
It mitigates the incorrect predictions generated by GLIP. 
Overall, CNVP leverages multi-view observations, which allows it to enhance prediction performance while retaining GLIP's zero-shot generalization.
\begin{figure*}[!tb]
  \includegraphics[width=\linewidth]{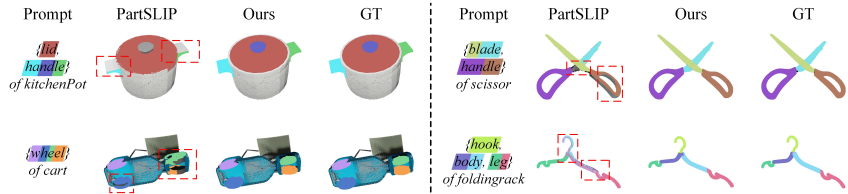}
\caption{Qualitative comparison on zero-shot instance segmentation (zoom in for details). 
Left: PartNetE's simulated data. Right: AKBSeg's real-world data. 
The red dashed boxes indicate that our method produces more accurate 3D segmentation boundaries compared to the SOTA method, PartSLIP.}
\label{fig:comparewithpartslip}
\end{figure*} 

\begin{table*}[!tb]
  \centering
  \caption{Zero-shot unlabeled segmentation results on the PartNetE benchmark. Object category Average IoUs(\%) are shown.}
  \resizebox{\linewidth}{!}{
\begin{tabular}{p{3.1cm}|c|cccccccccccc}
\toprule
Method & Overall (45) & Bottle & Door  & Lamp  & Scissors & Table & Box   & Kettle & KitchenPot & Lighter & Pliers & Stapler & Toilet \\
\midrule
PartSLIP \cite{liu2023partslip} & 36.4  & 78.0  & 27.9  & 47.6  & 47.2  & 46.9  & 38.8  & 66.2  & 60.5  & 53.2  & 3.5   & 27.3  & 45.4  \\
Ours (w/o Extending) & 45.6  & 55.7  & 23.9  & 42.8  & 39.9  & 47.9  & 51.6  & 57.8  & 70.7  & 47.3  & 40.1  & 39.7  & 56.0  \\
Ours  & \textbf{56.0} & \textbf{80.4} & \textbf{37.8} & \textbf{72.9} & \textbf{51.1} & \textbf{53.3} & \textbf{63.1} & \textbf{85.5} & \textbf{80.3} & \textbf{64.4} & \textbf{61.3} & \textbf{80.7} & \textbf{58.2} \\
\bottomrule
\multicolumn{12}{l}{Please refer to the supplementary material for the full table. This also applies to \cref{tab:unlabel_akbseg,tab:ins_partnete,tab:ins_akbseg}.} \\
\end{tabular}%
    }
  \label{tab:unlabel_partnete}
\end{table*}

\begin{table*}[!tb]
  \centering
  \caption{Instance segmentation results on the PartNetE benchmark. Object category mAP50s(\%) are shown.}
  \resizebox{\linewidth}{!}{
\begin{tabular}{p{3.1cm}|c|cccccccccccc}
\toprule
Method & Overall (45) & Bottle & Door  & Lamp  & Scissors & Table & Box   & Kettle & KitchenPot & Lighter & Pliers & Stapler & Toilet \\
\midrule
PointGroup\textsuperscript{*} \cite{jiang2020pointgroup} & 31.0  & 38.2  & \textbf{23.4} & 62.7  & 38.5  & \textbf{46.3} & 7.2   & 61.3  & \textbf{59.5} & 33.6  & 28.2  & 88.3  & 2.2  \\
SoftGroup\textsuperscript{*} \cite{vu2022softgroup} & \textbf{31.9} & \textbf{43.9} & 21.2  & \textbf{63.3} & \textbf{39.3} & 46.2  & \textbf{8.6} & \textbf{63.8} & 59.3  & \textbf{34.6} & \textbf{40.4} & \textbf{94.3} & \textbf{2.4} \\
\midrule
PartSLIP\textsuperscript{†} \cite{liu2023partslip} & 23.3  & 67.0  & 10.6  & 27.8  & 18.2  & 28.6  & 18.9  & 26.8  & 58.9  & 15.1  & 1.0   & 16.2  & 12.9  \\
Ours (w/o CNVP)\textsuperscript{†} & 24.1  & 62.0  & 14.8  & 30.9  & 26.1  & 26.6  & 26.0  & 22.0  & 54.4  & 16.3  & 38.6  & 22.3  & 14.3  \\
Ours\textsuperscript{†}  & \textbf{28.5} & \textbf{74.5} & \textbf{15.7} & \textbf{35.9} & \textbf{26.4} & \textbf{29.4} & \textbf{32.2} & \textbf{33.4} & \textbf{64.4} & \textbf{21.1} & \textbf{40.7} & \textbf{44.9} & \textbf{15.5} \\
\bottomrule
\multicolumn{12}{l}{* fully supervised; † zero-shot; PartSLIP's overall result reproduces by the official code, with the official paper being 18.0\% mAP50.}
\end{tabular}%
    }
  \label{tab:ins_partnete}
\end{table*} 

\begin{table*}[!tb]
  \centering
  \caption{Zero-shot unlabeled segmentation results on the AKBSeg benchmark. Object category Average IoUs(\%) are shown. }
  \resizebox{\linewidth}{!}{
\begin{tabular}{p{3.1cm}|c|cccccccccccc}
\toprule
Method & Overall (16) & Ballpoint & Bottle & Box   & Bucket & Condiment & Drink & Faucet & Foldingrack & Knife & Sauce & Shampoo & Trashcan \\
\midrule
PartSLIP \cite{liu2023partslip} & 34.3  & 3.0   & 8.7   & 35.0  & 49.9  & 44.4  & 10.5  & 11.0  & 32.4  & 73.0  & 22.5  & 37.6  & 66.0  \\
Ours (w/o Extending) & 49.3  & 33.7  & 55.7  & \textbf{54.1} & 74.1  & 50.7  & 49.9  & 44.6  & 50.6  & 74.8  & 35.7  & 56.5  & 72.0  \\
Ours  & \textbf{58.9} & \textbf{48.9} & \textbf{65.7} & 52.5  & \textbf{75.5} & \textbf{65.2} & \textbf{67.8} & \textbf{52.8} & \textbf{64.1} & \textbf{84.1} & \textbf{45.8} & \textbf{63.2} & \textbf{79.8} \\
\bottomrule
\end{tabular}%
    }
  \label{tab:unlabel_akbseg}
\end{table*} 

\begin{table*}[!tb]
  \centering
  \caption{Zero-shot instance segmentation results on the AKBSeg benchmark. Object category mAP50s(\%) are shown.}
  \resizebox{\linewidth}{!}{
\begin{tabular}{p{3.1cm}|c|cccccccccccc}
\toprule
Method & Overall (16) & Ballpoint & Bottle & Box   & Bucket & Condiment & Drink & Faucet & Foldingrack & Knife & Sauce & Shampoo & Trashcan \\
\midrule
PartSLIP \cite{liu2023partslip} & 15.0  & 1.0   & 1.1   & 12.9  & 34.8  & 11.4  & 1.0   & 3.4   & 16.7  & 51.0  & 5.1   & 2.8   & 9.3  \\
Ours (w/o CNVP) & 23.9  & 5.0   & \textbf{26.0} & 13.6  & 59.6  & 28.5  & 35.0  & \textbf{4.0} & 35.4  & 80.3  & 9.2   & 4.9   & 7.1  \\
Ours  & \textbf{26.5} & \textbf{6.5} & 20.2  & \textbf{16.3} & \textbf{77.8} & \textbf{41.0} & \textbf{36.8} & \textbf{4.0} & \textbf{36.2} & \textbf{80.9} & \textbf{10.1} & \textbf{6.6} & \textbf{10.0} \\
\bottomrule
\end{tabular}%
}
  \label{tab:ins_akbseg}
\end{table*} 

\section{Experiments} 
\subsection{Benchmark and Metric}
\textbf{PartNetE.} For PartNetE \cite{liu2023partslip}, the training data are 28,367 3D objects from PartNet \cite{mo2019partnet} and \(45 \times 8\) 3D objects from PartNet-Mobility \cite{xiang2020sapien}, and the testing data are 1906 3D objects from PartNet-Mobility covering 45 object categories. 
PartNetE encompasses both common coarse-grained (e.g., chair seat) and fine-grained (e.g., knob) parts. 
This diversity of levels of granularity presents a significant challenge for the evaluated method. 

\textbf{AKBSeg.} To better evaluate the generalization of all zero-shot baselines, we propose an AKBSeg benchmark. 
It collects 508 3D objects from the AKB-48 \cite{liu2022akb} dataset covering 16 object categories for testing data. Based on the original semantic annotations, we provide additional instance labels. 
Building upon PartNetE's simulated data, incorporating AKBSeg's real-world data into the experiment further enhances the zero-shot baseline's evaluative credibility. This also benefits the evaluation of future work in zero-shot 3D part segmentation. 

\textbf{Metric.} We follow \cite{wang2021learning} to utilize the Average IoU as the unlabeled segmentation metric. 
We use the mask of instance label as ground truth for evaluating the unlabeled segmentation. 
We follow \cite{liu2023partslip} to utilize the category mAP (50\% IoU threshold) as the instance segmentation metric. 

\subsection{Implementation Details}
\label{sec:Implementation Details} 
For our method, the input is an RGB object point cloud. 
The point cloud rendering resolution of PyTorch3D \cite{pytorch3D} is set to $800 \times 800$. 
The number of viewpoints is set to 20. For details of the viewpoints, please refer to the supplementary materials. 
The number of FPS output points in \cref{eq:KP2D} is set to $256$. 
Note that the FPS here should be distinguished from that in \cref{eq:KP2D1}. 
The merge threshold T is set to 0.3, and the ablation analysis is described in \cref{sec:ablationExtending}. 

\subsection{Comparison with Existing Methods} 
First, we conduct the quantitative evaluation on PartNetE with zero-shot methods and 3D fully supervised counterparts (See \cref{tab:unlabel_partnete,tab:ins_partnete}). 
Second, to further investigate the impact of factors concerning \textit{unseen data, unseen classes, and hyperparameters (which ablation studies cannot fully account for)}, we evaluate all zero-shot baselines on AKBSeg again (See \cref{tab:unlabel_akbseg,tab:ins_akbseg}). 
\textit{The same configuration is maintained for each baseline across PartNetE and AKBSeg.} 

\textbf{Zero-shot method.} 
We compare our method with the existing SOTA zero-shot instance segmentation method, PartSLIP. 
\cref{tab:unlabel_partnete,tab:unlabel_akbseg} show that, for zero-shot unlabeled segmentation, our method achieves 56.0\% and 58.9\% Average IoUs and outperforms PartSLIP by large margins. 
\cref{tab:ins_partnete,tab:ins_akbseg} show that, for zero-shot instance segmentation, our method outperforms PartSLIP by 5.2\% and 11.5\% mAP50s, respectively. 
From PartNetE to AKBSeg, our method nearly maintains its performance, while PartSLIP declines significantly. 
It indicates that our method demonstrates superior robustness. 
As shown in \cref{fig:comparewithpartslip}, since PartSLIP is based on superpoints \cite{landrieu2018large}, which are similar to superpixels in 2D, it is challenging to ensure accurate 3D segmentation boundaries. 
In contrast, our proposed self-extension lifts SAM's mask from 2D to 3D in a training-free manner, retaining zero-shot generalization while producing accurate 3D segmentation boundaries. 
Overall, our method significantly surpasses PartSLIP in both zero-shot generalization and segmentation performance. 

\textbf{3D fully supervised counterparts (methods).} To observe the gap between zero-shot and 3D fully supervised (3D-FS) methods, we evaluate all baselines on the PartNetE. 
The zero-shot methods do not use any 3D training data, while the 3D-FS methods are trained on \(45 \times 8+28k\) 3D objects. 
Since our method improves by 5.2\% mAP50 compared to PartSLIP, it reduces the maximum gap between zero-shot and 3D-FS methods from 8.6\% to 3.4\% mAP50. 
Moreover, notably zero-shot methods possess a unique advantage in predicting unseen classes, such as directly evaluating on AKBSeg (See \cref{tab:ins_partnete,tab:ins_akbseg}), unlike 3D-FS methods limited to predefined fixed classes before training. 

\begin{figure}[!tb]
     \centering
   \includegraphics[width=\linewidth]{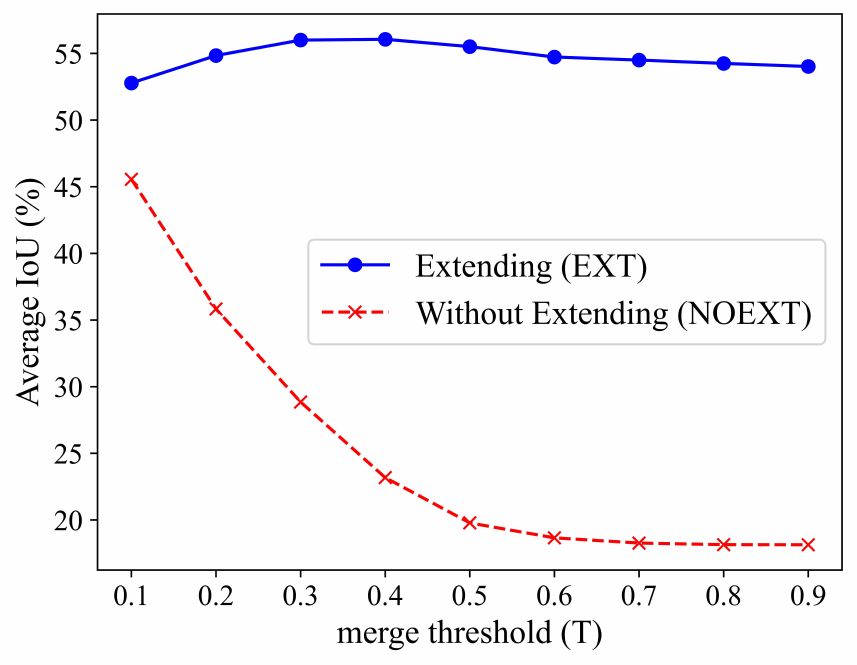}
   \caption{Ablation Study on self-extension by the `Extending' and `Without Extending' settings. The Average IoU is the overall result on PartNetE. See \cref{sec:ablationExtending} for details.}
   \label{fig:Tanalysis} 
\end{figure} 

\subsection{Ablation Study} 
\begin{figure}[!tb]
     \centering
   \includegraphics[width=\linewidth]{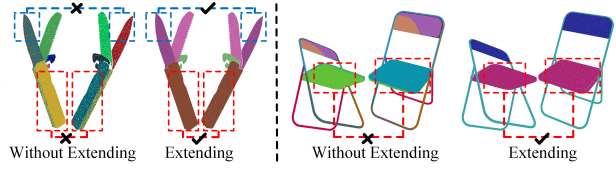}
   \caption{Qualitative results on self-extension (zoom in for details). See \cref{sec:ablationExtending} for details.}
   \label{fig:ablationExtending}
\end{figure}

\textbf{Self-extension.} 
\label{sec:ablationExtending} 
To analyze the effectiveness of the proposed self-extension, we conduct the ablation study by two settings (`Extending' and `Without Extending'). 
In the Extending (EXT) setting, we retain all self-extension processes. In the Without Extending (NOEXT) setting, we remove all steps involving continuously extending, namely all SVEs (See \cref{fig:self-extension}). In other words, we skip \cref{eq:SS2ES} in each self-extension. 
Through the results in \cref{fig:Tanalysis}, we observe the following: 1) NOEXT is overall lower than EXT. 
2) NOEXT is overly dependent on the threshold T. 
Although fixing T to 0.1 yields relatively good performance, this approach's robustness and stability are suboptimal. 
3) Compared to NOEXT, EXT demonstrates minimal sensitivity to changes in T, meanwhile consistently maintaining its superior performance. 
Overall, EXT demonstrates better robustness and stability compared to NOEXT. On the other hand, as shown in \cref{tab:unlabel_akbseg}, EXT similarly exhibits a significant performance improvement on the AKBSeg. Moreover, \cref{fig:ablationExtending} also shows that EXT produces better 3D consistency than NOEXT. 
This shows that self-extension effectively leverages the natural relationship between co-viewed regions and SAM’s prompt mechanism to lift 2D to 3D. 
Since EXT is almost independent of T, we set T to 0.3 in \cref{sec:Implementation Details}. 

\begin{table}[!tb]
  \centering
  \caption{Ablation study on the number of viewpoints. The Average IoU is the overall result on PartNetE.}
  \resizebox{.5\linewidth}{!}{
\begin{tabular}{c|c|c|c}
\toprule
viewpoints  & 20   & 8  & 4 \\
\midrule
Average IoU & 56.0  & 50.2  & 36.2  \\
\bottomrule
\end{tabular}%
}
\label{tab:numberofviewpoints}
\end{table} 

\textbf{Class Non-highest Vote Penalty (CNVP).} 
We conduct the ablation study on the proposed CNVP. 
As shown in \cref{tab:ins_partnete,tab:ins_akbseg}, CNVP consistently maintains performance gains across the PartNetE and AKBSeg benchmarks, with improvements of 4.4\% and 2.6\% mAP50s, respectively. 
This indicates that CNVP effectively utilizes multi-view observations, enhancing performance while retaining GLIP's zero-shot generalization. 

\textbf{Number of Viewpoints.} 
We conduct the ablation study on the number of viewpoints. As shown in \cref{tab:numberofviewpoints}, when the number drops to 8, the high performance is still maintained. 
Since it is difficult for 4 viewpoints to cover the entire 3D object uniformly, the performance drops substantially.

\section{Conclusion} 
In this work, we propose a novel zero-shot 3D part segmentation pipeline. 
We explore the natural relationship between multi-view correspondence and the FMs' prompt mechanisms. The relationship manifests in the pipeline as self-extension, TDCM, and CNVP. 
Through extensive qualitative and quantitative comparison and ablation studies, our method demonstrates superior zero-shot generalization and segmentation performance than the SOTA method. 

\vspace{1em} 
\noindent\textbf{Acknowledgements.} We thank Minghua Liu for providing the code and participating in discussions. This work was supported by National Natural Science Foundation of China under Grant No. 61702340; and in part by the Start-up Foundation for Introducing Talent of Nanjing University of Information Science \& Technology (NUIST) under Grant 2023r063.

\clearpage
\setcounter{table}{0}
\renewcommand{\thetable}{S\arabic{table}}
\renewcommand{\thesection}{S} 
\setcounter{figure}{0}
\renewcommand{\thefigure}{S\arabic{figure}} 

\section{Supplementary Material}
\subsection{Merging 3D Groups} 
The core steps of the merging algorithm are presented in the \cref{alg:Algorithm1}. 
Starting from $K_1$ different viewpoints, we obtain a set of $m_{1}$ 3D groups from all self-extensions, denoted as $A=\{G_{1}^{3D},\ldots,G_{m_{1}}^{3D}\}$. 
Since $G^{3D}$ with the same semantics in set $A$ have similar areas, we sort set $A$ in descending order. 
All sets in this algorithm are regarded as ordered sets. 
Then we iterate over $A$, and for each $G^{3D}$, it is either merged with an existing $M^{3D}$ in $B$ or added to $B$ as a new $M^{3D}$ (steps 3-12). 
Note that we use the Intersection over Union (IoU) as the criterion to determine whether $G^{3D}$ is merged or not, while controlled by the merge threshold T. 
Second, we need to ensure that each point of $Q^{3D}$ is associated with unique semantics. 
If a point simultaneously exists in different $M^{3D}$, we choose to assign it to the $M^{3D}$ with higher granularity. 
Since $M^{3D}$ in $B$ is granularity from lower to higher, we iterate over $B$ and add $M^{3D}$ to $C$ each time. 
After adding $M^{3D}$ each time, it is needed to remove the points that each $P^{3D}$ in $C$ (except the currently added $M^{3D}$) shares with the current $M^{3D}$ (step 14-16). 
Finally, we return set $C=\{P_{1}^{3D},\ldots,P_{m_{2}}^{3D}\}$ which includes $m_{2}$ 3D unlabeled parts $P^{3D}$. 

\subsection{Implementation Details of Viewpoints} 
\begin{figure}[!b]
    \centering
   \includegraphics[width=.8\linewidth]{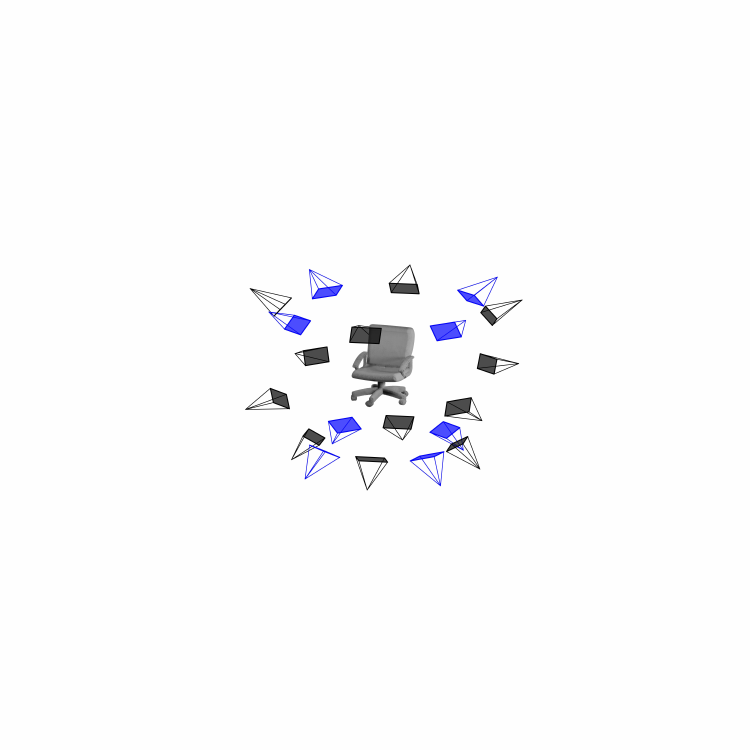}
   \caption{Visualization of viewpoint positions. 
   In the experiment, we place 20 viewpoints around the 3D object, with 8 of these viewpoints (highlighted in blue) each as a starting viewpoint.}
   \label{fig:viewpointposition}
\end{figure}

In the experiment, we follow PartSLIP \cite{liu2023partslip} to fix the viewpoint position and set the camera distance to 2.2 in Pytorch3D. 
We place 20 viewpoints around the 3D object, with 8 of these viewpoints each serving as a starting viewpoint. The viewpoint positions are detailed in \cref{tab:vposition,fig:viewpointposition}. 

\begin{algorithm}[!t]
\caption{Merge 3D Groups. T is the merge threshold.}
\label{alg:Algorithm1}
\renewcommand{\algorithmicrequire}{\textbf{Input:}}
\renewcommand{\algorithmicensure}{\textbf{Output:}}
\newcommand{\Break}{\State \textbf{break}}
\newcommand{\Continue}{\State \textbf{continue}}
\begin{algorithmic}[1]
        \Require
                3D groups $A=\{G_{1}^{3D},\ldots,G_{m_{1}}^{3D}\}$   
        \Ensure 3D parts $C=\{P_{1}^{3D},\ldots,P_{m_{2}}^{3D}\}$    
        \State sort the elements in $A$ by area (the number of points) in descending order
        \State initialize an empty set $B$
        \For{each $G^{3D}$ \textbf{in} $A$}
            \State $flag = False$
            \For{each $M^{3D}$ \textbf{in} $B$}
                \State calculate $iou$ of $G^{3D}$ and $M^{3D}$
                \If {$iou > T$}
                    \State $M^{3D} \leftarrow  M^{3D} \cup G^{3D}$ \Comment {update $M^{3D}$ in $B$}
                    \State $flag = True$
                    \Break
                \EndIf
            \EndFor
            \If{\textbf{not} $flag$}
               \State add $G^{3D}$ to $B$ 
            \EndIf
        \EndFor
        \State initialize an empty set $C$
        \For{each $M^{3D}$ \textbf{in} $B$}
            \State add $M^{3D}$ to $C$
            \State $C[0 : \text{len}(C) - 1] \leftarrow C[0 : \text{len}(C) - 1] \setminus M^{3D}$ 
        \EndFor
        \State \Return $C$
\end{algorithmic}
\end{algorithm} 

\begin{table}[!t]
\centering
\caption{List of all viewpoints with elevation and azimuth angles.}
  \resizebox{.6\linewidth}{!}{
\begin{tabular}{l|c|c}
\toprule
id & elevation (°) & azimuth (°) \\
\midrule
1\textsuperscript{*}  & 35  & -35  \\
2  & 35  & 10   \\
3\textsuperscript{*}  & 35  & 55   \\
4  & 35  & 100  \\
5\textsuperscript{*}  & 35  & 145  \\
6  & 35  & 190  \\
7\textsuperscript{*}  & 35  & 235  \\
8  & 35  & 280  \\
9  & -10 & -35  \\
10 & -10 & 55   \\
11 & -10 & 145  \\
12 & -10 & 235  \\
13\textsuperscript{*} & -55 & -35  \\
14 & -55 & 10   \\
15\textsuperscript{*} & -55 & 55   \\
16 & -55 & 100  \\
17\textsuperscript{*} & -55 & 145  \\
18 & -55 & 190  \\
19\textsuperscript{*} & -55 & 235  \\
20 & -55 & 280  \\
\bottomrule
\multicolumn{3}{l}{* starting viewpoint} \\
\end{tabular}
}
\label{tab:vposition}
\end{table} 

\subsection{AKBSeg Benchmark}
\cref{tab:akbsegStatistics} shows the statistics of the proposed AKBSeg benchmark. 
All 508 3D objects collect from the AKB-48 \cite{liu2022akb} dataset. 
To be consistent with PartNetE's requirements for the part instance segmentation, we provide additional instance labels for the original AKB-48 data. 
While the PartNetE and AKBSeg benchmarks have overlapping object categories, the 3D parts exhibit significant differences. 
For example, in the object category `Trashcan', PartNetE includes `footpedal', `lid', and `door', whereas AKBSeg includes `lid' and `wheel'. 
Such differences and diversity are advantageous for evaluating zero-shot baselines, as they are entirely grounded in real-world scenarios. 
We hope that the proposed AKBSeg benchmark could help the future evaluation of zero-shot 3D part segmentation. 

\begin{table}[!t]
  \centering
  \caption{The table shows the statistics of the AKBSeg benchmark.} 
  \resizebox{.6\linewidth}{!}{
\begin{tabular}{ccc}
\toprule
category & parts & test \\
\midrule
Ballpoint & cap,button & 9 \\
Bottle & lid   & 35 \\
Box   & lid   & 40 \\
Bucket & handle & 37 \\
Condiment & lid,handle & 10 \\
Cup   & lid,handle & 34 \\
Drink & lid   & 51 \\
Eyeglasses & body,leg & 93 \\
Faucet & spout,switch & 45 \\
Foldingrack & hook,body,leg & 11 \\
Knife & blade,handle & 9 \\
Lighter & lid,wheel,button & 19 \\
Sauce & lid   & 43 \\
Scissor & blade,handle & 19 \\
Shampoo & head,lid & 31 \\
Trashcan & lid,wheel & 22 \\
\midrule
16 in total & 29 in total & 508 \\
\bottomrule
\end{tabular}%
 }
  \label{tab:akbsegStatistics}
\end{table} 

\subsection{More Quantitative Comparison} 
\textbf{Zero-shot Semantic Segmentation.} We degrade the instance segmentation result into semantic segmentation and compare it with existing methods. 
We follow \cite{liu2023partslip} to utilize the category mIoU as the metric. 
For PointCLIP V2's \cite{zhu2023pointclip} text prompt, we follow it to prompt GPT-3 \cite{brown2020language} to generate 3D specific text for each part category of the input object by constructing a 3D language command. 
\cref{tab:zero-shot_semantic} shows that our method on the semantic segmentation task consistently exhibits performance advantages over other methods, similar to the gap observed on the unlabeled and instance segmentation tasks. 

\begin{table}[!t]
  \centering
  \caption{Zero-shot semantic segmentation results, measured as mIoU(\%).} 
  \resizebox{.7\linewidth}{!}{
\begin{tabular}{cccc}
\toprule
      & PointCLIP V2 \cite{zhu2023pointclip}  & PartSLIP \cite{liu2023partslip} & Ours \\
\midrule
PartNetE & 16.1  & 34.4  & \textbf{39.3} \\
AKBSeg & 17.8  & 25.9  & \textbf{35.7} \\
\bottomrule
\end{tabular}%
}
\label{tab:zero-shot_semantic}
\end{table} 

\textbf{Comparison with PartSLIP++.} We conduct the quantitative comparison with PartSLIP++ \cite{zhou2023partslip++}, \textit{a concurrent work} that introduces two main improvements to PartSLIP: 1) using SAM to refine the GLIP's bounding boxes, to achieve more accurate 2D predictions; 2) proposing an improved Expectation-Maximization (EM) algorithm, to lift 2D segmentation to 3D. 
Since the improved EM algorithm is based on the few-shot setting, we replaced it with the PartSLIP's voting strategy to ensure the zero-shot setting. 
As shown in \cref{tab:partnetE_partslip++,tab:AKBSeg_partslip++}, our method demonstrates better zero-shot performance compared to PartSLIP++ across the different tasks and benchmarks. 

\begin{table}[!t]
  \centering
  \caption{Quantitative comparison with PartSLIP++ on PartNetE.} 
  \resizebox{.8\linewidth}{!}{
\begin{tabular}{cccc}
\toprule
      & unlabeled seg.  & instance seg. & semantic seg. \\
\midrule
PartSLIP++ \cite{zhou2023partslip++}  & 38.9  & 25.5  & 37.4  \\
Ours  & \textbf{56.0} & \textbf{28.5} & \textbf{39.3} \\
\bottomrule
\end{tabular}%
}
\label{tab:partnetE_partslip++}
\end{table} 

\begin{table}[!t]
  \centering
  \caption{Quantitative comparison with PartSLIP++ on AKBSeg.} 
  \resizebox{.8\linewidth}{!}{
\begin{tabular}{cccc}
\toprule
      & unlabeled seg. & instance seg. & semantic seg. \\
\midrule
PartSLIP++ \cite{zhou2023partslip++} & 35.7  & 17.5  & 26.8  \\
Ours  & \textbf{58.9} & \textbf{26.5} & \textbf{35.7} \\
\bottomrule
\end{tabular}%
}
\label{tab:AKBSeg_partslip++}
\end{table} 

\begin{table}[!t]
  \centering
  \caption{Ablation Study on TDCM, measured as mAP50(\%).} 
  \resizebox{.5\linewidth}{!}{

\begin{tabular}{cccc}
\toprule
      & 2D    & 3D    & 2D \& 3D \\
\midrule
PartNetE & 22.8  & 19.5  & \textbf{24.1} \\
\midrule
AKBSeg & 22.6  & 20.1  & \textbf{23.9} \\
\bottomrule
\end{tabular}%
}
\label{tab:ablation_TDCM}
\end{table} 

\begin{figure}[!t]
     \centering
   \includegraphics[width=\linewidth]{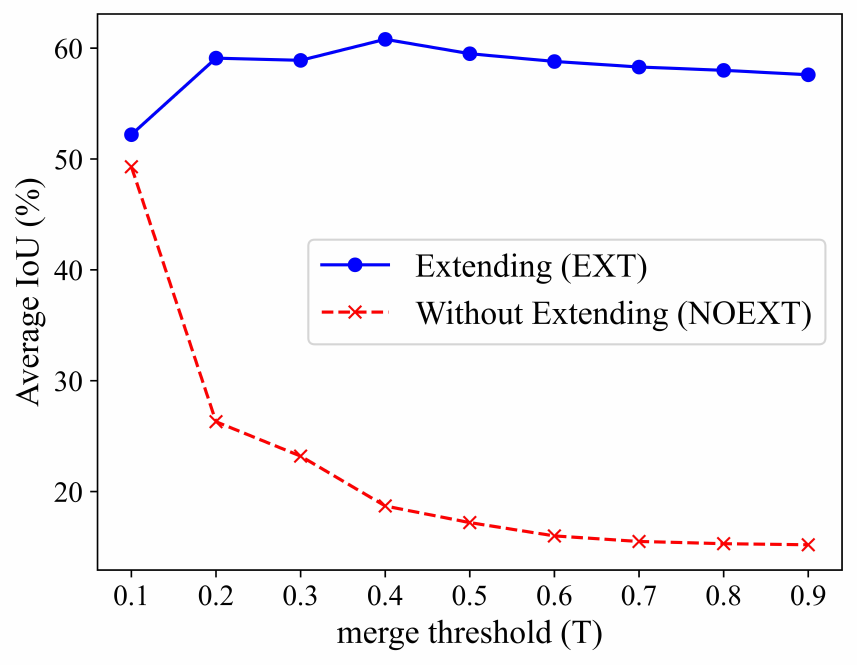}
   \caption{Ablation Study on self-extension by the `Extending' and `Without Extending' settings. The Average IoU is the overall result on AKBSeg.}
   \label{fig:T_AKBSeg}
\end{figure} 

\subsection{More Ablation Studies}
\textbf{Self-extension.} 
To further analyze the effectiveness of self-extension, we conduct the ablation study on the AKBSeg benchmark. 
As shown in \cref{fig:T_AKBSeg}, from PartNetE (See Fig.6 in main paper) to AKBSeg, EXT and NOEXT exhibit similar performance trends. Compared to NOEXT, EXT still demonstrates better robustness and stability. 

\textbf{Two-dimensional Checking Mechanism (TDCM).} 
To evaluate the effectiveness of TDCM, we conduct the ablation study. 
As shown in \cref{tab:ablation_TDCM}, when voting is performed only in either the 2D or 3D space, we observe the performance decrease. 
This indicates that aligning the voting results from both 2D and 3D spaces, to check and then discard unqualified bounding boxes is effective and reasonable. 

\begin{figure}[!t]
     \centering
   \includegraphics[width=\linewidth]{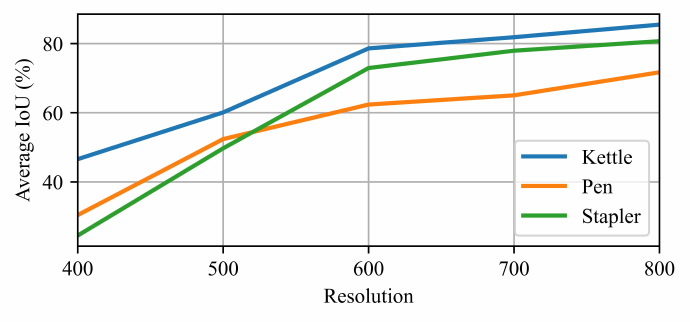}
   \caption{Ablation study on render resolution.}
   \label{fig:reso}
\end{figure}

\textbf{Render Resolution.} We conduct the ablation study on the rendering resolution. As shown in \cref{fig:reso}, increasing the resolution from 400 to 600 results in a significant performance gain. However, the gain is relatively small when increasing the resolution from 600 to 800. 

\begin{table}[!t]
  \centering
  \caption{Sensitivities to Object Rotation. Applying random rotation to the input object results in no significant fluctuation in performance, measured as Average IoU(\%).} 
  \resizebox{.5\linewidth}{!}{
\begin{tabular}{c|ccc}
\toprule
setting & Kettle & Pen   & Stapler \\
\midrule
original & \textbf{85.5} & 71.7  & 80.7 \\
rotated & 85.2  & \textbf{72.1} & \textbf{80.9} \\
\bottomrule
\end{tabular}%
}
\label{tab:rotated}
\end{table} 

\textbf{Sensitivities to Object Rotation.} We perform sensitivity analysis on the input object by random rotations. 
As shown in \cref{tab:rotated}, we observe no significant fluctuation in performance. 
This indicates that our method is hardly sensitive to object rotation. 

\subsection{Metric Details}
For the unlabeled segmentation, we follow \cite{wang2021learning} to utilize the Average IoU as its metric. 
For each 3D object, the Average IoU is determined by calculating the maximum IoU for each ground-truth instance part from the predicted parts and averaging them.  Next, we calculate the average metrics for each object category. 
For the instance segmentation, We follow \cite{liu2023partslip} to utilize mAP (50\% IoU threshold) as its metrics. 
For each object category, the mAP50\% is determined by calculating the AP50\% of each part instance category and averaging them. 

\subsection{Text Prompts} 
The official Supplementary Material and code of PartSLIP demonstrate that the text prompt without the object name gives GLIP better performance (\eg, removing `of chair' in `arm, back, leg, seat, wheel of chair'). 
Thus, our method and PartSLIPs' text prompts contain only the part names (\eg, `arm, back, leg, seat, wheel') in the experiment. 

\subsection{Density of Input Points} 
To obtain finer 3D parts, we follow \cite{liu2023partslip,wang2021learning} to focus on the dense rather than sparse point clouds in this work. 

\subsection{Discussion and Limitation} 
While the current pipeline demonstrates strong zero-shot generalization and segmentation performance, it has certain limitations. The primary limitation is that the performance of the pretrained foundation models directly impacts our pipeline. However, since our pipeline relies only on the foundation models' prompt mechanism, the proposed three training-free manners (self-extension, TDCM, and CNVP) are independent of the internal structure of the foundation models. Therefore, replacing the pretrained foundation model with another model that uses the same prompt mechanism is a reasonable solution. Another generally applicable approach is to fine-tune the foundation model. 

\subsection{Full Table of Quantitative Comparison}
\cref{tab:full_un_partnetE,tab:full_ins_partnetE,tab:full_un_akbseg,tab:full_ins_akbseg} show the full tables of quantitative comparison results of the main paper.

\begin{table}[!t]
  \centering
  \caption{Full table of zero-shot unlabeled segmentation results on the PartNetE benchmark, corresponding to Table 1 of the main paper. Object category Average IoUs(\%) are shown.}
  \resizebox{.85\linewidth}{!}{
\begin{tabular}{c|ccc}
\toprule
category & PartSLIP \cite{liu2023partslip} & Ours (w/o Extending) & Ours \\
\midrule
Bottle & 78.0  & 55.7  & \textbf{80.4} \\
Chair & \textbf{76.8} & 60.1  & 71.8  \\
Clock & 17.8  & \textbf{38.1} & 33.8  \\
Dishwasher & 37.3  & \textbf{63.3} & 59.5  \\
Display & \textbf{73.5} & 56.9  & 68.7  \\
Door  & 27.9  & 23.9  & \textbf{37.8} \\
Faucet & 16.5  & 45.0  & \textbf{75.5} \\
Keyboard & 0.5   & \textbf{29.6} & 27.7  \\
Knife & 18.3  & 44.5  & \textbf{68.7} \\
Lamp  & 47.6  & 42.8  & \textbf{72.9} \\
Laptop & 41.2  & \textbf{47.3} & 39.8  \\
Microwave & 14.1  & \textbf{14.6} & 13.0  \\
Refrigerator & 32.2  & \textbf{61.3} & 57.9  \\
Scissors & 47.2  & 39.9  & \textbf{51.1} \\
StorageFurniture & 49.8  & \textbf{55.8} & 50.0  \\
Table & 46.9  & 47.9  & \textbf{53.3} \\
TrashCan & 51.1  & 58.1  & \textbf{67.2} \\
Box   & 38.8  & 51.6  & \textbf{63.1} \\
Bucket & 35.6  & 57.3  & \textbf{83.7} \\
Camera & 36.7  & \textbf{48.1} & 40.9  \\
Cart  & \textbf{78.7} & 58.2  & 78.5  \\
CoffeeMachine & 28.9  & 29.2  & \textbf{31.1} \\
Dispenser & 40.6  & 52.4  & \textbf{74.6} \\
Eyeglasses & 5.5   & 36.7  & \textbf{70.4} \\
FoldingChair & \textbf{85.5} & 44.9  & 76.2  \\
Globe & \textbf{89.5} & 33.4  & 53.0  \\
Kettle & 66.2  & 57.8  & \textbf{85.5} \\
KitchenPot & 60.5  & 70.7  & \textbf{80.3} \\
Lighter & 53.2  & 47.3  & \textbf{64.4} \\
Mouse & 21.7  & \textbf{38.9} & 34.9  \\
Oven  & 22.8  & \textbf{45.5} & 37.0  \\
Pen   & 44.6  & 56.1  & \textbf{71.7} \\
Phone & 11.8  & \textbf{56.7} & 38.4  \\
Pliers & 3.5   & 40.1  & \textbf{61.3} \\
Printer & 1.1   & \textbf{6.2} & 4.1  \\
Remote & 2.8   & \textbf{38.0} & 34.3  \\
Safe  & 17.4  & \textbf{26.5} & 26.2  \\
Stapler & 27.3  & 39.7  & \textbf{80.7} \\
Suitcase & \textbf{65.8} & 51.8  & 62.2  \\
Switch & 6.5   & 68.0  & \textbf{72.4} \\
Toaster & 19.7  & \textbf{65.0} & 62.7  \\
Toilet & 45.4  & 56.0  & \textbf{58.2} \\
USB   & 35.3  & 32.4  & \textbf{85.0} \\
WashingMachine & 14.8  & \textbf{16.9} & 15.6  \\
Window & 2.5   & 39.8  & \textbf{46.7} \\
Overall (45) & 36.4  & 45.6  & \textbf{56.0} \\
\bottomrule
\end{tabular}%
    
}
\label{tab:full_un_partnetE}
\end{table}

\begin{table}[!t]
  \centering
  \caption{Full table of zero-shot unlabeled segmentation results on the AKBSeg benchmark, corresponding to Table 3 of the main paper. Object category Average IoUs(\%) are shown.}
  \resizebox{.85\linewidth}{!}{
  \begin{tabular}{c|ccc}
\toprule
category & PartSLIP \cite{liu2023partslip} & Ours (w/o Extending) & Ours \\
\midrule
Ballpoint & 3.0   & 33.7  & \textbf{48.9} \\
Bottle & 8.7   & 55.7  & \textbf{65.7} \\
Box   & 35.0  & \textbf{54.1} & 52.5  \\
Bucket & 49.9  & 74.1  & \textbf{75.5} \\
Condiment & 44.4  & 50.7  & \textbf{65.2} \\
Cup   & \textbf{42.2} & 28.2  & 39.5  \\
Drink & 10.5  & 49.9  & \textbf{67.8} \\
Eyeglasses & 1.0   & 25.0  & \textbf{38.2} \\
Faucet & 11.0  & 44.6  & \textbf{52.8} \\
Foldingrack & 32.4  & 50.6  & \textbf{64.1} \\
Knife & 73.0  & 74.8  & \textbf{84.1} \\
Lighter & \textbf{39.1} & 34.9  & 33.0  \\
Sauce & 22.5  & 35.7  & \textbf{45.8} \\
Scissor & \textbf{73.1} & 47.8  & 66.9  \\
Shampoo & 37.6  & 56.5  & \textbf{63.2} \\
Trashcan & 66.0  & 72.0  & \textbf{79.8} \\
Overall (16) & 34.3  & 49.3  & \textbf{58.9} \\
\bottomrule
\end{tabular}%
}
    
    \label{tab:full_un_akbseg}
\end{table}

\begin{table}[!t]
  \centering
  \caption{Full table of zero-shot instance segmentation results on the AKBSeg benchmark, corresponding to Table 4 of the main paper. Object category mAP50s(\%) are shown.}
  \resizebox{.85\linewidth}{!}{
  \begin{tabular}{cc|ccc}
\toprule
category & parts & PartSLIP \cite{liu2023partslip} & Ours (w/o CNVP)  & Ours \\
\midrule
\multirow{2}[2]{*}{Ballpoint} & cap   & \textbf{1.0} & \textbf{1.0} & \textbf{1.0} \\
      & button & 1.0   & 8.9   & \textbf{12.1} \\
\midrule
Bottle & lid   & 1.1   & \textbf{26.0} & 20.2  \\
\midrule
Box   & lid   & 12.9  & 13.6  & \textbf{16.3} \\
\midrule
Bucket & handle & 34.8  & 59.6  & \textbf{77.8} \\
\midrule
\multirow{2}[2]{*}{Condiment} & lid   & 7.9   & 13.9  & \textbf{21.6} \\
      & handle & 14.9  & 43.1  & \textbf{60.4} \\
\midrule
\multirow{2}[2]{*}{Cup} & lid   & 1.1   & 4.7   & \textbf{5.2} \\
      & handle & \textbf{42.3} & 16.4  & 24.4  \\
\midrule
Drink & lid   & 1.0   & 35.0  & \textbf{36.8} \\
\midrule
\multirow{2}[2]{*}{Eyeglasses} & body  & 1.0   & \textbf{5.4} & 4.0  \\
      & leg   & 1.0   & \textbf{12.1} & 10.3  \\
\midrule
\multirow{2}[2]{*}{Faucet} & spout & 1.0   & 1.7   & \textbf{3.3} \\
      & switch & 5.8   & \textbf{6.3} & 4.6  \\
\midrule
\multirow{3}[2]{*}{Foldingrack} & hook  & 48.0  & 61.5  & \textbf{76.6} \\
      & body  & 1.0   & \textbf{22.7} & 20.0  \\
      & leg   & 1.0   & \textbf{22.0} & 12.0  \\
\midrule
\multirow{2}[2]{*}{Knife} & blade & 65.9  & 72.5  & \textbf{73.6} \\
      & handle & 36.1  & \textbf{88.1} & \textbf{88.1} \\
\midrule
\multirow{3}[2]{*}{Lighter} & lid   & \textbf{1.0} & \textbf{1.0} & \textbf{1.0} \\
      & wheel & \textbf{1.0} & \textbf{1.0} & \textbf{1.0} \\
      & button & \textbf{6.1} & 1.0   & 1.0  \\
\midrule
Sauce & lid   & 5.1   & 9.2   & \textbf{10.1} \\
\midrule
\multirow{2}[2]{*}{Scissor} & blade & \textbf{41.9} & 38.9  & 38.1  \\
      & handle & \textbf{87.3} & 68.5  & 71.1  \\
\midrule
\multirow{2}[2]{*}{Shampoo} & head  & 1.5   & 7.1   & \textbf{10.9} \\
      & lid   & \textbf{4.2} & 2.7   & 2.3  \\
\midrule
\multirow{2}[2]{*}{Trashcan} & lid   & 1.8   & 4.6   & \textbf{4.6} \\
      & wheel & \textbf{16.8} & 9.6   & 15.4  \\
\midrule
\multicolumn{2}{c|}{Overall (16)} & 15.0  & 23.9  & \textbf{26.5} \\
\bottomrule
\end{tabular}%
} 
\label{tab:full_ins_akbseg} 
\end{table}  

\begin{table*}[!t]
  \caption{Full table of instance segmentation results on the PartNetE benchmark, corresponding to Table 2 of the main paper. Object category mAP50s(\%) are shown.}
  \resizebox{\linewidth}{!}{
\begin{tabular}{ccccccccccccccc}
\toprule
category & \multicolumn{1}{c|}{part} & PointGroup\textsuperscript{*} \cite{jiang2020pointgroup}& \multicolumn{1}{c|}{SoftGroup\textsuperscript{*} \cite{vu2022softgroup}} & PartSLIP\textsuperscript{†} \cite{liu2023partslip} & Ours (w/o CNVP)\textsuperscript{†} & \multicolumn{1}{c|}{Ours\textsuperscript{†}} & \multicolumn{1}{c|}{} & category & \multicolumn{1}{c|}{part} & PointGroup\textsuperscript{*} \cite{jiang2020pointgroup}& \multicolumn{1}{c|}{SoftGroup\textsuperscript{*} \cite{vu2022softgroup}} & PartSLIP\textsuperscript{†} \cite{liu2023partslip} & Ours (w/o CNVP)\textsuperscript{†} & Ours\textsuperscript{†} \\
\midrule
Bottle & \multicolumn{1}{c|}{lid} & 38.2  & \multicolumn{1}{c|}{\textbf{43.9}} & 67.0  & 62.0  & \multicolumn{1}{c|}{\textbf{74.5}} & \multicolumn{1}{c|}{} & \multirow{2}[4]{*}{Camera} & \multicolumn{1}{c|}{button} & 1.0   & \multicolumn{1}{c|}{\textbf{1.5}} & \textbf{19.9} & 10.4  & 8.5  \\
\cmidrule{1-7}\multirow{5}[6]{*}{Chair} & \multicolumn{1}{c|}{arm} & 94.6  & \multicolumn{1}{c|}{\textbf{95.1}} & 44.4  & 47.0  & \multicolumn{1}{c|}{\textbf{67.3}} & \multicolumn{1}{c|}{} &       & \multicolumn{1}{c|}{lens} & \textbf{16.1} & \multicolumn{1}{c|}{0.0} & \textbf{23.0} & 21.0  & 19.8  \\
\cmidrule{9-15}      & \multicolumn{1}{c|}{back} & \textbf{82.0} & \multicolumn{1}{c|}{73.2} & \textbf{86.4} & 67.1  & \multicolumn{1}{c|}{72.0} & \multicolumn{1}{c|}{} & Cart  & \multicolumn{1}{c|}{wheel} & \textbf{29.2} & \multicolumn{1}{c|}{28.4} & 80.7  & 83.0  & \textbf{86.7} \\
\cmidrule{9-15}      & \multicolumn{1}{c|}{leg} & 88.6  & \multicolumn{1}{c|}{\textbf{93.6}} & \textbf{52.3} & 46.4  & \multicolumn{1}{c|}{52.1} & \multicolumn{1}{c|}{} & \multirow{4}[4]{*}{CoffeeMachine} & \multicolumn{1}{c|}{button} & \textbf{1.0} & \multicolumn{1}{c|}{\textbf{1.0}} & \textbf{6.9} & 3.1   & 1.6  \\
      & \multicolumn{1}{c|}{seat} & 75.0  & \multicolumn{1}{c|}{\textbf{85.9}} & \textbf{87.2} & 50.7  & \multicolumn{1}{c|}{69.4} & \multicolumn{1}{c|}{} &       & \multicolumn{1}{c|}{container} & 2.5   & \multicolumn{1}{c|}{\textbf{4.0}} & 16.8  & 14.3  & \textbf{17.4} \\
      & \multicolumn{1}{c|}{wheel} & \textbf{98.0} & \multicolumn{1}{c|}{97.7} & \textbf{92.6} & 58.4  & \multicolumn{1}{c|}{67.2} & \multicolumn{1}{c|}{} &       & \multicolumn{1}{c|}{knob} & \textbf{5.6} & \multicolumn{1}{c|}{5.0} & \textbf{7.5} & 4.9   & 5.8  \\
\cmidrule{1-7}Clock & \multicolumn{1}{c|}{hand} & \textbf{1.0} & \multicolumn{1}{c|}{\textbf{1.0}} & 3.1   & \textbf{10.9} & \multicolumn{1}{c|}{9.5} & \multicolumn{1}{c|}{} &       & \multicolumn{1}{c|}{lid} & \textbf{3.3} & \multicolumn{1}{c|}{1.4} & 11.3  & 14.2  & \textbf{22.2} \\
\cmidrule{1-7}\cmidrule{9-15}\multirow{2}[2]{*}{Dishwasher} & \multicolumn{1}{c|}{door} & \textbf{76.7} & \multicolumn{1}{c|}{75.0} & 13.4  & 13.1  & \multicolumn{1}{c|}{\textbf{21.1}} & \multicolumn{1}{c|}{} & \multirow{2}[2]{*}{Dispenser} & \multicolumn{1}{c|}{head} & 27.5  & \multicolumn{1}{c|}{\textbf{29.2}} & 1.8   & \textbf{2.8} & \textbf{2.8} \\
      & \multicolumn{1}{c|}{handle} & 55.6  & \multicolumn{1}{c|}{\textbf{56.4}} & 16.1  & 22.9  & \multicolumn{1}{c|}{\textbf{32.3}} & \multicolumn{1}{c|}{} &       & \multicolumn{1}{c|}{lid} & 20.5  & \multicolumn{1}{c|}{\textbf{23.6}} & 8.2   & 8.7   & \textbf{12.1} \\
\cmidrule{1-7}\cmidrule{9-15}\multirow{3}[4]{*}{Display} & \multicolumn{1}{c|}{base} & 95.2  & \multicolumn{1}{c|}{\textbf{97.4}} & 71.1  & 51.8  & \multicolumn{1}{c|}{\textbf{72.8}} & \multicolumn{1}{c|}{} & \multirow{2}[2]{*}{Eyeglasses} & \multicolumn{1}{c|}{body} & 31.7  & \multicolumn{1}{c|}{\textbf{39.5}} & 4.2   & \textbf{9.9} & 8.0  \\
      & \multicolumn{1}{c|}{screen} & 46.0  & \multicolumn{1}{c|}{\textbf{55.4}} & 25.5  & 30.9  & \multicolumn{1}{c|}{\textbf{32.5}} & \multicolumn{1}{c|}{} &       & \multicolumn{1}{c|}{leg} & \textbf{68.0} & \multicolumn{1}{c|}{62.7} & 1.0   & \textbf{60.3} & 56.1  \\
\cmidrule{9-15}      & \multicolumn{1}{c|}{support} & \textbf{54.0} & \multicolumn{1}{c|}{53.2} & 38.0  & 30.4  & \multicolumn{1}{c|}{\textbf{45.1}} & \multicolumn{1}{c|}{} & FoldingChair & \multicolumn{1}{c|}{seat} & \textbf{16.8} & \multicolumn{1}{c|}{\textbf{16.8}} & \textbf{83.3} & 70.3  & 75.0  \\
\cmidrule{1-7}\cmidrule{9-15}\multirow{3}[4]{*}{Door} & \multicolumn{1}{c|}{frame} & \textbf{36.8} & \multicolumn{1}{c|}{28.3} & 1.0   & \textbf{22.0} & \multicolumn{1}{c|}{19.2} & \multicolumn{1}{c|}{} & Globe & \multicolumn{1}{c|}{sphere} & \textbf{63.1} & \multicolumn{1}{c|}{\textbf{63.1}} & \textbf{90.9} & 17.6  & 25.5  \\
\cmidrule{9-15}      & \multicolumn{1}{c|}{door} & 32.4  & \multicolumn{1}{c|}{\textbf{34.3}} & 15.4  & 21.1  & \multicolumn{1}{c|}{\textbf{26.4}} & \multicolumn{1}{c|}{} & \multirow{3}[4]{*}{Kettle} & \multicolumn{1}{c|}{lid} & 64.0  & \multicolumn{1}{c|}{\textbf{64.4}} & 28.8  & 25.8  & \textbf{40.2} \\
      & \multicolumn{1}{c|}{handle} & \textbf{1.0} & \multicolumn{1}{c|}{\textbf{1.0}} & \textbf{15.3} & 1.3   & \multicolumn{1}{c|}{1.4} & \multicolumn{1}{c|}{} &       & \multicolumn{1}{c|}{handle} & 51.4  & \multicolumn{1}{c|}{\textbf{54.3}} & 50.4  & 39.1  & \textbf{55.9} \\
\cmidrule{1-7}\multirow{2}[4]{*}{Faucet} & \multicolumn{1}{c|}{spout} & 85.4  & \multicolumn{1}{c|}{\textbf{86.3}} & 8.0   & 21.2  & \multicolumn{1}{c|}{\textbf{23.3}} & \multicolumn{1}{c|}{} &       & \multicolumn{1}{c|}{spout} & 68.5  & \multicolumn{1}{c|}{\textbf{72.6}} & 1.0   & 1.0   & \textbf{4.3} \\
\cmidrule{9-15}      & \multicolumn{1}{c|}{switch} & \textbf{74.5} & \multicolumn{1}{c|}{72.5} & 2.3   & \textbf{15.3} & \multicolumn{1}{c|}{12.7} & \multicolumn{1}{c|}{} & \multirow{2}[4]{*}{KitchenPot} & \multicolumn{1}{c|}{lid} & 68.3  & \multicolumn{1}{c|}{\textbf{68.5}} & \textbf{85.2} & 50.0  & 65.7  \\
\cmidrule{1-7}\multirow{2}[4]{*}{Keyboard} & \multicolumn{1}{c|}{cord} & \textbf{42.6} & \multicolumn{1}{c|}{39.7} & \textbf{75.4} & 38.3  & \multicolumn{1}{c|}{60.1} & \multicolumn{1}{c|}{} &       & \multicolumn{1}{c|}{handle} & \textbf{50.6} & \multicolumn{1}{c|}{50.1} & 32.7  & 58.8  & \textbf{63.1} \\
\cmidrule{9-15}      & \multicolumn{1}{c|}{key} & 37.2  & \multicolumn{1}{c|}{\textbf{37.7}} & 1.0   & \textbf{4.3} & \multicolumn{1}{c|}{2.6} & \multicolumn{1}{c|}{} & \multirow{3}[6]{*}{Lighter} & \multicolumn{1}{c|}{lid} & \textbf{30.7} & \multicolumn{1}{c|}{\textbf{30.7}} & 30.7  & 20.1  & \textbf{32.5} \\
\cmidrule{1-7}Knife & \multicolumn{1}{c|}{blade} & 19.3  & \multicolumn{1}{c|}{\textbf{27.2}} & 11.6  & \textbf{28.0} & \multicolumn{1}{c|}{21.5} & \multicolumn{1}{c|}{} &       & \multicolumn{1}{c|}{wheel} & \textbf{6.0} & \multicolumn{1}{c|}{5.3} & 9.0   & 9.4   & \textbf{12.4} \\
\cmidrule{1-7}\multirow{4}[4]{*}{Lamp} & \multicolumn{1}{c|}{base} & 64.3  & \multicolumn{1}{c|}{\textbf{71.1}} & \textbf{75.6} & 59.7  & \multicolumn{1}{c|}{72.2} & \multicolumn{1}{c|}{} &       & \multicolumn{1}{c|}{button} & 64.1  & \multicolumn{1}{c|}{\textbf{67.8}} & 5.5   & \textbf{19.3} & 18.5  \\
\cmidrule{9-15}      & \multicolumn{1}{c|}{body} & \textbf{48.6} & \multicolumn{1}{c|}{36.5} & 1.4   & 18.1  & \multicolumn{1}{c|}{\textbf{20.3}} & \multicolumn{1}{c|}{} & \multirow{3}[2]{*}{Mouse} & \multicolumn{1}{c|}{button} & \textbf{1.0} & \multicolumn{1}{c|}{\textbf{1.0}} & 6.3   & \textbf{8.8} & 7.7  \\
      & \multicolumn{1}{c|}{bulb} & 54.5  & \multicolumn{1}{c|}{\textbf{59.2}} & 1.4   & \textbf{3.8} & \multicolumn{1}{c|}{2.2} & \multicolumn{1}{c|}{} &       & \multicolumn{1}{c|}{cord} & \textbf{1.0} & \multicolumn{1}{c|}{\textbf{1.0}} & \textbf{55.4} & 27.1  & 27.1  \\
      & \multicolumn{1}{c|}{shade} & 83.5  & \multicolumn{1}{c|}{\textbf{86.4}} & 32.7  & 41.8  & \multicolumn{1}{c|}{\textbf{48.7}} & \multicolumn{1}{c|}{} &       & \multicolumn{1}{c|}{wheel} & \textbf{83.2} & \multicolumn{1}{c|}{\textbf{83.2}} & 35.3  & 46.3  & \textbf{50.5} \\
\cmidrule{1-7}\cmidrule{9-15}\multirow{5}[6]{*}{Laptop} & \multicolumn{1}{c|}{keyboard} & \textbf{0.0} & \multicolumn{1}{c|}{\textbf{0.0}} & \textbf{34.5} & 5.9   & \multicolumn{1}{c|}{7.1} & \multicolumn{1}{c|}{} & \multirow{2}[2]{*}{Oven} & \multicolumn{1}{c|}{door} & 26.5  & \multicolumn{1}{c|}{\textbf{31.9}} & \textbf{42.8} & 24.5  & 23.4  \\
      & \multicolumn{1}{c|}{screen} & \textbf{1.0} & \multicolumn{1}{c|}{\textbf{1.0}} & 34.5  & 41.6  & \multicolumn{1}{c|}{\textbf{50.6}} & \multicolumn{1}{c|}{} &       & \multicolumn{1}{c|}{knob} & \textbf{1.0} & \multicolumn{1}{c|}{\textbf{1.0}} & 7.9   & \textbf{21.7} & 19.0  \\
\cmidrule{9-15}      & \multicolumn{1}{c|}{shaft} & 1.2   & \multicolumn{1}{c|}{\textbf{3.5}} & \textbf{1.0} & \textbf{1.0} & \multicolumn{1}{c|}{\textbf{1.0}} & \multicolumn{1}{c|}{} & \multirow{2}[2]{*}{Pen} & \multicolumn{1}{c|}{cap} & \textbf{48.2} & \multicolumn{1}{c|}{44.4} & 2.3   & 5.5   & \textbf{10.5} \\
      & \multicolumn{1}{c|}{touchpad} & \textbf{0.0} & \multicolumn{1}{c|}{\textbf{0.0}} & 14.5  & 30.1  & \multicolumn{1}{c|}{\textbf{35.7}} & \multicolumn{1}{c|}{} &       & \multicolumn{1}{c|}{button} & \textbf{16.9} & \multicolumn{1}{c|}{\textbf{16.9}} & 1.0   & \textbf{1.1} & \textbf{1.1} \\
\cmidrule{9-15}      & \multicolumn{1}{c|}{camera} & \textbf{0.0} & \multicolumn{1}{c|}{\textbf{0.0}} & \textbf{1.0} & \textbf{1.0} & \multicolumn{1}{c|}{\textbf{1.0}} & \multicolumn{1}{c|}{} & \multirow{2}[4]{*}{Phone} & \multicolumn{1}{c|}{lid} & 1.0   & \multicolumn{1}{c|}{\textbf{1.1}} & 13.4  & 51.3  & \textbf{55.4} \\
\cmidrule{1-7}\multirow{4}[8]{*}{Microwave} & \multicolumn{1}{c|}{display} & \textbf{4.2} & \multicolumn{1}{c|}{1.0} & \textbf{22.8} & \textbf{22.8} & \multicolumn{1}{c|}{\textbf{22.8}} & \multicolumn{1}{c|}{} &       & \multicolumn{1}{c|}{button} & \textbf{1.0} & \multicolumn{1}{c|}{\textbf{1.0}} & 8.8   & \textbf{19.7} & 6.2  \\
\cmidrule{9-15}      & \multicolumn{1}{c|}{door} & \textbf{62.6} & \multicolumn{1}{c|}{57.1} & 12.6  & 18.0  & \multicolumn{1}{c|}{\textbf{35.1}} & \multicolumn{1}{c|}{} & Pliers & \multicolumn{1}{c|}{leg} & 28.2  & \multicolumn{1}{c|}{\textbf{40.4}} & 1.0   & 38.6  & \textbf{40.7} \\
\cmidrule{9-15}      & \multicolumn{1}{c|}{handle} & \textbf{1.0} & \multicolumn{1}{c|}{\textbf{1.0}} & \textbf{100.0} & 14.6  & \multicolumn{1}{c|}{16.4} & \multicolumn{1}{c|}{} & Printer & \multicolumn{1}{c|}{button} & \textbf{1.0} & \multicolumn{1}{c|}{\textbf{1.0}} & \textbf{1.3} & 1.0   & 1.0  \\
\cmidrule{9-15}      & \multicolumn{1}{c|}{button} & \textbf{100.0} & \multicolumn{1}{c|}{\textbf{100.0}} & \textbf{4.1} & 2.2   & \multicolumn{1}{c|}{1.9} & \multicolumn{1}{c|}{} & Remote & \multicolumn{1}{c|}{button} & \textbf{23.4} & \multicolumn{1}{c|}{22.5} & 3.0   & \textbf{5.2} & 2.7  \\
\cmidrule{1-7}\cmidrule{9-15}\multirow{2}[2]{*}{Refrigerator} & \multicolumn{1}{c|}{door} & \textbf{57.1} & \multicolumn{1}{c|}{54.2} & 16.9  & 15.8  & \multicolumn{1}{c|}{\textbf{17.7}} & \multicolumn{1}{c|}{} & \multirow{3}[4]{*}{Safe} & \multicolumn{1}{c|}{door} & 11.0  & \multicolumn{1}{c|}{\textbf{12.3}} & 4.4   & \textbf{10.9} & 10.7  \\
      & \multicolumn{1}{c|}{handle} & \textbf{19.3} & \multicolumn{1}{c|}{17.2} & 23.0  & 26.2  & \multicolumn{1}{c|}{\textbf{33.3}} & \multicolumn{1}{c|}{} &       & \multicolumn{1}{c|}{switch} & 4.8   & \multicolumn{1}{c|}{\textbf{5.4}} & 1.5   & 3.5   & \textbf{4.6} \\
\cmidrule{1-7}\multirow{3}[4]{*}{Scissors} & \multicolumn{1}{c|}{blade} & 6.2   & \multicolumn{1}{c|}{\textbf{6.5}} & 5.4   & \textbf{7.0} & \multicolumn{1}{c|}{\textbf{7.0}} & \multicolumn{1}{c|}{} &       & \multicolumn{1}{c|}{button} & \textbf{1.0} & \multicolumn{1}{c|}{\textbf{1.0}} & \textbf{1.0} & \textbf{1.0} & \textbf{1.0} \\
\cmidrule{9-15}      & \multicolumn{1}{c|}{handle} & 82.0  & \multicolumn{1}{c|}{\textbf{82.9}} & 40.1  & 67.0  & \multicolumn{1}{c|}{\textbf{67.7}} & \multicolumn{1}{c|}{} & \multirow{2}[2]{*}{Stapler} & \multicolumn{1}{c|}{body} & 86.6  & \multicolumn{1}{c|}{\textbf{96.7}} & \textbf{0.0} & \textbf{0.0} & \textbf{0.0} \\
      & \multicolumn{1}{c|}{screw} & 27.2  & \multicolumn{1}{c|}{\textbf{28.4}} & \textbf{9.0} & 4.3   & \multicolumn{1}{c|}{4.6} & \multicolumn{1}{c|}{} &       & \multicolumn{1}{c|}{lid} & 90.0  & \multicolumn{1}{c|}{\textbf{91.8}} & 32.4  & 44.7  & \textbf{89.9} \\
\cmidrule{1-7}\cmidrule{9-15}\multirow{3}[4]{*}{StorageFurniture} & \multicolumn{1}{c|}{door} & \textbf{86.9} & \multicolumn{1}{c|}{85.6} & \textbf{10.2} & 4.7   & \multicolumn{1}{c|}{7.5} & \multicolumn{1}{c|}{} & \multirow{2}[2]{*}{Suitcase} & \multicolumn{1}{c|}{handle} & \textbf{25.5} & \multicolumn{1}{c|}{24.2} & 35.2  & 50.9  & \textbf{71.4} \\
      & \multicolumn{1}{c|}{drawer} & 3.9   & \multicolumn{1}{c|}{\textbf{4.2}} & \textbf{7.9} & 5.2   & \multicolumn{1}{c|}{4.9} & \multicolumn{1}{c|}{} &       & \multicolumn{1}{c|}{wheel} & \textbf{5.7} & \multicolumn{1}{c|}{2.9} & 17.7  & 18.7  & \textbf{24.5} \\
\cmidrule{9-15}      & \multicolumn{1}{c|}{handle} & 56.4  & \multicolumn{1}{c|}{\textbf{57.5}} & 33.0  & 25.3  & \multicolumn{1}{c|}{\textbf{33.6}} & \multicolumn{1}{c|}{} & Switch & \multicolumn{1}{c|}{switch} & \textbf{7.5} & \multicolumn{1}{c|}{5.6} & 3.2   & \textbf{4.6} & 4.4  \\
\cmidrule{1-7}\cmidrule{9-15}\multirow{6}[6]{*}{Table} & \multicolumn{1}{c|}{door} & 44.4  & \multicolumn{1}{c|}{\textbf{49.3}} & 5.7   & \textbf{5.9} & \multicolumn{1}{c|}{5.1} & \multicolumn{1}{c|}{} & \multirow{2}[2]{*}{Toaster} & \multicolumn{1}{c|}{button} & 9.0   & \multicolumn{1}{c|}{\textbf{10.1}} & 13.8  & 12.9  & \textbf{13.9} \\
      & \multicolumn{1}{c|}{drawer} & 35.7  & \multicolumn{1}{c|}{\textbf{36.5}} & 7.4   & 6.9   & \multicolumn{1}{c|}{\textbf{7.9}} & \multicolumn{1}{c|}{} &       & \multicolumn{1}{c|}{slider} & \textbf{5.0} & \multicolumn{1}{c|}{\textbf{5.0}} & \textbf{0.0} & \textbf{0.0} & \textbf{0.0} \\
\cmidrule{9-15}      & \multicolumn{1}{c|}{leg} & \textbf{33.8} & \multicolumn{1}{c|}{27.4} & 24.4  & 35.4  & \multicolumn{1}{c|}{\textbf{43.9}} & \multicolumn{1}{c|}{} & \multirow{3}[2]{*}{Toilet} & \multicolumn{1}{c|}{lid} & 5.5   & \multicolumn{1}{c|}{\textbf{6.1}} & 23.7  & 31.0  & \textbf{36.2} \\
      & \multicolumn{1}{c|}{tabletop} & 81.2  & \multicolumn{1}{c|}{\textbf{82.0}} & 47.4  & 63.8  & \multicolumn{1}{c|}{\textbf{73.2}} & \multicolumn{1}{c|}{} &       & \multicolumn{1}{c|}{seat} & \textbf{0.0} & \multicolumn{1}{c|}{\textbf{0.0}} & 2.4   & \textbf{3.5} & 2.6  \\
      & \multicolumn{1}{c|}{wheel} & 1.0   & \multicolumn{1}{c|}{\textbf{1.3}} & \textbf{75.4} & 45.0  & \multicolumn{1}{c|}{43.3} & \multicolumn{1}{c|}{} &       & \multicolumn{1}{c|}{button} & \textbf{1.0} & \multicolumn{1}{c|}{\textbf{1.0}} & \textbf{12.5} & 8.2   & 7.6  \\
\cmidrule{9-15}      & \multicolumn{1}{c|}{handle} & \textbf{81.9} & \multicolumn{1}{c|}{80.8} & \textbf{11.1} & 2.4   & \multicolumn{1}{c|}{3.1} & \multicolumn{1}{c|}{} & \multirow{2}[4]{*}{USB} & \multicolumn{1}{c|}{cap} & 67.3  & \multicolumn{1}{c|}{\textbf{75.7}} & 14.6  & 20.4  & \textbf{25.5} \\
\cmidrule{1-7}\multirow{3}[4]{*}{TrashCan} & \multicolumn{1}{c|}{footpedal} & 34.8  & \multicolumn{1}{c|}{\textbf{35.3}} & 0.0   & \textbf{1.0} & \multicolumn{1}{c|}{\textbf{1.0}} & \multicolumn{1}{c|}{} &       & \multicolumn{1}{c|}{rotation} & \textbf{16.3} & \multicolumn{1}{c|}{15.0} & 1.0   & 1.0   & \textbf{17.6} \\
\cmidrule{9-15}      & \multicolumn{1}{c|}{lid} & \textbf{0.0} & \multicolumn{1}{c|}{\textbf{0.0}} & 40.3  & 17.1  & \multicolumn{1}{c|}{\textbf{45.0}} & \multicolumn{1}{c|}{} & \multirow{2}[2]{*}{WashingMachine} & \multicolumn{1}{c|}{door} & 25.0  & \multicolumn{1}{c|}{\textbf{34.3}} & \textbf{48.8} & 27.4  & 38.8  \\
      & \multicolumn{1}{c|}{door} & \textbf{0.0} & \multicolumn{1}{c|}{\textbf{0.0}} & \textbf{7.7} & 2.5   & \multicolumn{1}{c|}{2.4} & \multicolumn{1}{c|}{} &       & \multicolumn{1}{c|}{button} & \textbf{0.0} & \multicolumn{1}{c|}{\textbf{0.0}} & \textbf{4.1} & 2.4   & 3.2  \\
\cmidrule{1-7}\cmidrule{9-15}Box   & \multicolumn{1}{c|}{lid} & 7.2   & \multicolumn{1}{c|}{\textbf{8.6}} & 18.9  & 26.0  & \multicolumn{1}{c|}{\textbf{32.2}} & \multicolumn{1}{c|}{} & Window & \multicolumn{1}{c|}{window} & 21.2  & \multicolumn{1}{c|}{\textbf{26.4}} & 1.0   & 4.0   & \textbf{4.5} \\
\cmidrule{1-7}\cmidrule{9-15}Bucket & \multicolumn{1}{c|}{handle} & 1.5   & \multicolumn{1}{c|}{\textbf{1.6}} & 8.8   & 60.0  & \multicolumn{1}{c|}{\textbf{75.6}} & \multicolumn{1}{c|}{} & \multicolumn{2}{c|}{Overall (45)} & 31.0  & \multicolumn{1}{c|}{\textbf{31.9}} & 23.3  & 24.1  & \textbf{28.5} \\
\midrule
\multicolumn{15}{l}{* fully supervised; † zero-shot; PartSLIP's overall result reproduces by the official code, with the official paper being 18.0\% mAP50.} \\
\end{tabular}%
}
    \label{tab:full_ins_partnetE}
\end{table*}

{
    \small
    \bibliographystyle{ieeenat_fullname}
    \bibliography{main}
}
\end{document}